\documentclass[]{jingdong}

\usepackage{amsmath,amssymb}
\usepackage{algorithm}
\usepackage{algorithmic}
\usepackage{newfloat}
\usepackage{listings}
\DeclareCaptionStyle{ruled}{labelfont=normalfont,labelsep=colon,strut=off}
\floatstyle{ruled}
\newfloat{listing}{tb}{lst}{}
\floatname{listing}{Listing}
\usepackage{array}
\usepackage{wrapfig}

\def\UrlFont{\rm}
\microtypesetup{expansion=false}
\title{ComboShoppingBench: Evaluating LLM Agents for Budget-Constrained Basket Shopping with Coupons}
\author{Adrian Li}
\author{Kelong Mao}
\author{Yudong Guo}
\author{Heming Xia}
\author{Xinwei Yang}
\author{\protect\\[+0.25em]Lirui Luo}
\author{Jace Wong}
\author{Pu Yao}
\author{Sulong Xu}
\author{Simiu Gu}

\affiliation{JD.com}

\abstract{Real-world shopping often requires constructing a basket of complementary items rather than retrieving a single product. Such combo-shopping tasks arise in device setup, meal preparation, event planning, and group takeout ordering, requiring joint reasoning about item compatibility, availability, store-level requirements, delivery fees, coupons, and budgets. Evaluation is challenging because multiple baskets may satisfy the same request, making exact-match metrics unsuitable, whereas semantic evaluation alone cannot detect infeasible orders, invalid coupon combinations, or incorrect payments. We introduce \textsc{ComboShoppingBench}, an agentic shopping benchmark for open-ended yet verifiable basket construction in a simulated commerce and takeout environment. During task synthesis, an exploration agent constructs a feasible and semantically coherent basket of purchasable products; this witness guides the generation of coupons, budget constraints, user queries, and aligned evaluation rubrics. During evaluation, LLM judges assess semantic satisfaction, response quality, and claim faithfulness, while deterministic validation checks product-ID validity, budget compliance, and coupon optimality. Experiments with diverse LLM agents demonstrate that even strong agents struggle on \textsc{ComboShoppingBench}, highlighting substantial room for improvement in reliable, constraint-aware combo shopping.}

\begin{document}

\maketitle

\addtocontents{toc}{\protect\setcounter{tocdepth}{-1}}
\section{Introduction}

Large language models (LLMs) are increasingly being developed as tool-using
shopping agents that can search product catalogs, inspect attributes,
compare alternatives, and execute purchasing decisions under user-specified
constraints. This evolution moves shopping assistance beyond single-query
product retrieval toward multi-step, goal-oriented decision making, where
agents must reason, act, and verify outcomes through interaction with
external environments.

\begin{figure*}[t]
\centering
\includegraphics[width=\textwidth]{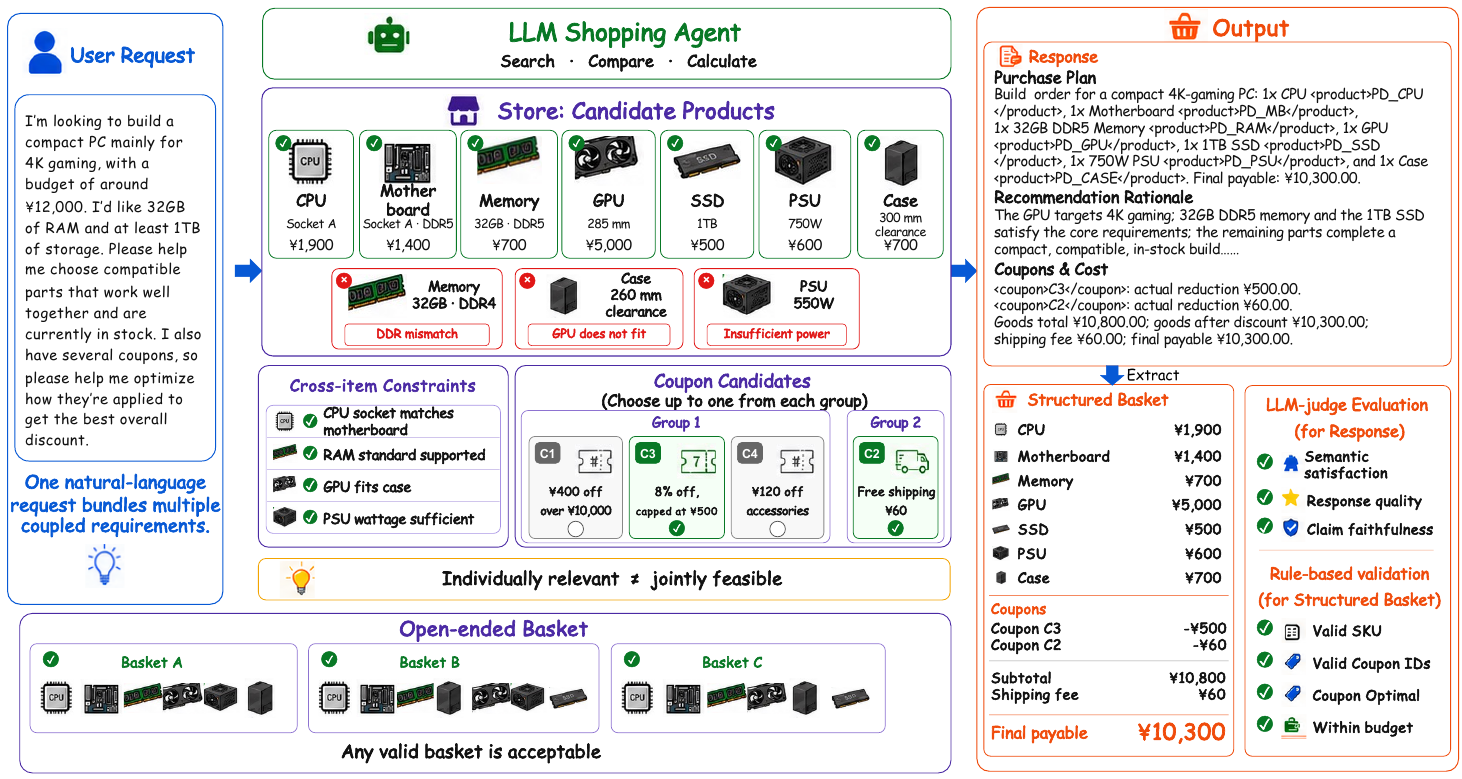}
\caption{Illustration of a combo-shopping task. The agent must select compatible items, construct executable orders, choose a legal coupon combination, and meet a final-payable budget. Multiple baskets may be valid, so evaluation verifies the agent's proposed basket.}
\label{fig:intro}
\end{figure*}

A particularly challenging setting emerges when a user goal requires
selecting and coordinating multiple products rather than retrieving
a single relevant item. We refer to this setting as \textbf{combo shopping}.
Unlike conventional recommendation tasks, combo shopping requires an agent to
construct a basket whose items are jointly compatible, operationally
feasible, and economically valid. For example, as shown in Figure~\ref{fig:intro}, building a compact PC for
4K gaming under a fixed budget requires selecting components that not only
satisfy individual preferences but also jointly meet constraints such as
socket compatibility, supported memory standards, power requirements, and
physical dimensions. Furthermore, the final transaction must respect
real-world purchasing constraints, including product availability, coupon
eligibility, and budget limits. Therefore, combo shopping represents an
open-ended decision problem in which multiple valid solutions may exist, but
each solution must satisfy a complex set of constraints.

Building a benchmark for combo shopping introduces two fundamental challenges.
First, task synthesis must generate natural and diverse user requests with meaningful cross-item dependencies while ensuring that every synthesized task admits at least one feasible basket. This is challenging because constraints that are individually reasonable may render the synthesized task unsolvable. Second, evaluation must reliably verify arbitrary agent-generated baskets. Since a task may admit multiple valid baskets, it is impractical to enumerate all correct solutions in advance. The key challenge is therefore to determine whether any generated basket constitutes a valid solution.

To this end, we introduce \textsc{ComboShoppingBench}, an interactive benchmark that formulates combo shopping as a basket-level agent task across e-commerce, takeout or instant retail, and mixed-domain scenarios. Beyond evaluating whether an agent constructs a suitable basket, \textsc{ComboShoppingBench} provides a comprehensive assessment of the complete shopping process, including order executability, settlement-based budget compliance, coupon legality and optimality, recommendation rationale quality, and claim faithfulness. Its central design principle is to construct tasks from a solution without evaluating against that solution: an exploration agent first identifies and validates a purchasable basket in the environment, which is then retained as a hidden witness for synthesizing the user request, budget, and transactional constraints. This solution-first construction ensures that every task is solvable by design, while the witness is never used as a reference answer during evaluation. An evaluated agent may return any basket that independently constitutes a valid solution to the task.

To evaluate each agent-proposed solution on its own merits, \textsc{ComboShoppingBench} adopts a four-dimensional framework that assesses both the proposed basket and the final response. Semantic satisfaction evaluates whether the selected products fulfill the user’s shopping intent. Rule-based validation first reconstructs a structured basket from the purchase plan described in the agent’s response, and then verifies the validity of the selected SKUs and coupons, budget compliance, and the optimality of coupon usage. Response quality assesses how effectively the solution is communicated, while claim faithfulness checks whether the factual statements in the response are consistent with the environment-recomputed results.

\begin{table*}[t]
\caption{Comparison of representative shopping-agent benchmarks.
Yes denotes explicit and complete support; Partial denotes a restricted
or indirect form of the capability; No denotes that the capability is
not evaluated.}
\label{tab:benchmark-comparison}
\centering
\setlength{\tabcolsep}{3.5pt}
\renewcommand{\arraystretch}{1.12}

\small
\resizebox{\textwidth}{!}{%
\begin{tabular}{@{}lcccccc@{}}
\toprule
Benchmark
& \shortstack{Basket-Level\\Task}
& \shortstack{Executable\\Order}
& \shortstack{Settlement-Based\\Budget}
& \shortstack{Coupon Legality\\\& Optimality}
& \shortstack{Recommendation\\Rationale Quality}
& \shortstack{Claim\\Faithfulness} \\
\midrule

WebShop
\cite{yao2023webshopscalablerealworldweb}
& No & Partial & No & No & No & No \\

DeepShop
\cite{lyu2025deepshopbenchmarkdeepresearch}
& No & No & No & No & No & No \\

ShoppingComp
\cite{tou2026shoppingcompllmsreallyready}
& No & No & No & No & Yes & No \\

WebMall
\cite{peeters2026webmallmultishopbenchmark}
& Partial & Partial & No & No & No & No \\

ShoppingBench
\cite{wang2026shoppingbenchrealworldintentgroundedshopping}
& Partial & Partial & Partial & Partial & No & No \\

Shopping Companion
\cite{yu2026shoppingcompanionbenchmarkingtraining}
& Yes & No & Partial & Partial & No & No \\

EComAgentBench
\cite{du2026ecomagentbenchbenchmarkingshoppingagents}
& No & No & Partial & Partial & No & No \\

\textbf{ComboShoppingBench}
& \textbf{Yes}
& \textbf{Yes}
& \textbf{Yes}
& \textbf{Yes}
& \textbf{Yes}
& \textbf{Yes} \\

\bottomrule
\end{tabular}
}
\end{table*}

We evaluate 11 proprietary and open-source agents under Think and No-think configurations, yielding 22 configurations. The strongest agent, GPT-5.5 with thinking enabled, achieves an Overall success rate of only 61.2\%. Failure analyses reveal persistent limitations in compositional requirement satisfaction, coupon and budget optimization, and faithful reporting, demonstrating that current shopping agents remain far from reliable end-to-end combo shopping.

Our main contributions are as follows:
\begin{itemize}
\item We formulate combo shopping as an agentic basket-construction problem in which an agent must identify and coordinate multiple complementary items under joint semantic and transactional constraints.

\item We introduce \textsc{ComboShoppingBench}, a benchmark spanning e-commerce, takeout or instant retail, and mixed-domain shopping. It features a solution-first construction pipeline that guarantees task feasibility through validated hidden witness baskets and a four-dimensional hybrid evaluation framework that assesses semantic satisfaction, rule-based validation, response quality, and claim faithfulness.

\item We comprehensively evaluate 11 agents. Even the best agent achieves an overall success rate of only 61.2\%, highlighting the substantial challenges posed by \textsc{ComboShoppingBench}. Quantitative and qualitative analyses further reveal persistent limitations in satisfying compositional requirements, adhering to coupon and budget constraints, and making faithful claims.
\end{itemize}

\section{Related Work}
\paragraph{Web shopping agent benchmarks.}
Web shopping agent benchmarks have evolved from browser-action grounding toward increasingly realistic evaluations of shopping decision making. Early work primarily tests whether agents can translate natural-language requests into sequences of search, navigation, option-selection, and checkout actions in web environments
\cite{yao2023webshopscalablerealworldweb,
deng2023mind2webgeneralistagentweb,
zhou2024webarenarealisticwebenvironment,
koh2024visualwebarenaevaluatingmultimodalagents,
lu2024weblinx}.
Shopping-specific benchmarks subsequently broaden the scope to richer attribute constraints, comparison and research, multi-store search, budget-aware purchasing, and long-horizon assistance involving preference learning, clarification, and proactive support
\cite{lyu2025deepshopbenchmarkdeepresearch,
tou2026shoppingcompllmsreallyready,
peeters2026webmallmultishopbenchmark,
wang2026shoppingbenchrealworldintentgroundedshopping,
wang2026shopsimulatorevaluatingexploringrldriven,
savadikar2026shopgymintegratedframeworkrealistic,
yu2026shoppingcompanionbenchmarkingtraining,
du2026ecomagentbenchbenchmarkingshoppingagents}. However, as summarized in Table~\ref{tab:benchmark-comparison}, existing shopping-agent benchmarks primarily focus on the retrieval or purchase of individual products, while those incorporating basket-level settings assess only limited aspects of the task. In contrast, \textsc{ComboShoppingBench} provides a comprehensive evaluation of the entire combo-shopping process, including response quality, claim faithfulness, and transaction validity.

\paragraph{Basket and bundle recommendation.}
Basket and bundle recommendation methods model item relationships, user preferences, and purchasing behavior to suggest coherent item sets. Prior work spans compatibility learning from behavioral and content signals
\cite{DBLP:conf/kdd/McAuleyPL15}, preference modeling over user--item--bundle interactions
\cite{DBLP:conf/sigir/PathakGM17,
DBLP:conf/ijcai/ChenL0GZ19,
DBLP:conf/sigir/ChangG0JL20,
DBLP:conf/sigir/LiLWSXYYZL21}, and intent-aware, contrastive, dynamic, and cold-start bundle generation
\cite{DBLP:conf/aaai/Zhao0ZM22,
DBLP:conf/kdd/MaHZWC22,
DBLP:journals/tors/SunFYFQOL24,
sun2024bundlesurvey,
bui2024bridge,
zhang2025residualdiffusion,
li2026epiccbr}. Next-basket recommendation instead predicts future purchase sets from transaction histories
\cite{cao2026case,deng2026timeintervalnbr}. These methods are typically evaluated as offline recommendation or prediction, without requiring interactive catalog search, cross-item constraint verification, store-policy compliance, or an executable purchase plan. \textsc{ComboShoppingBench} instead evaluates basket construction as an interactive agent task combining product compatibility with transaction-level constraint satisfaction.

\section{ComboShoppingBench}

\begin{figure*}[t]
\centering
\includegraphics[width=\textwidth]{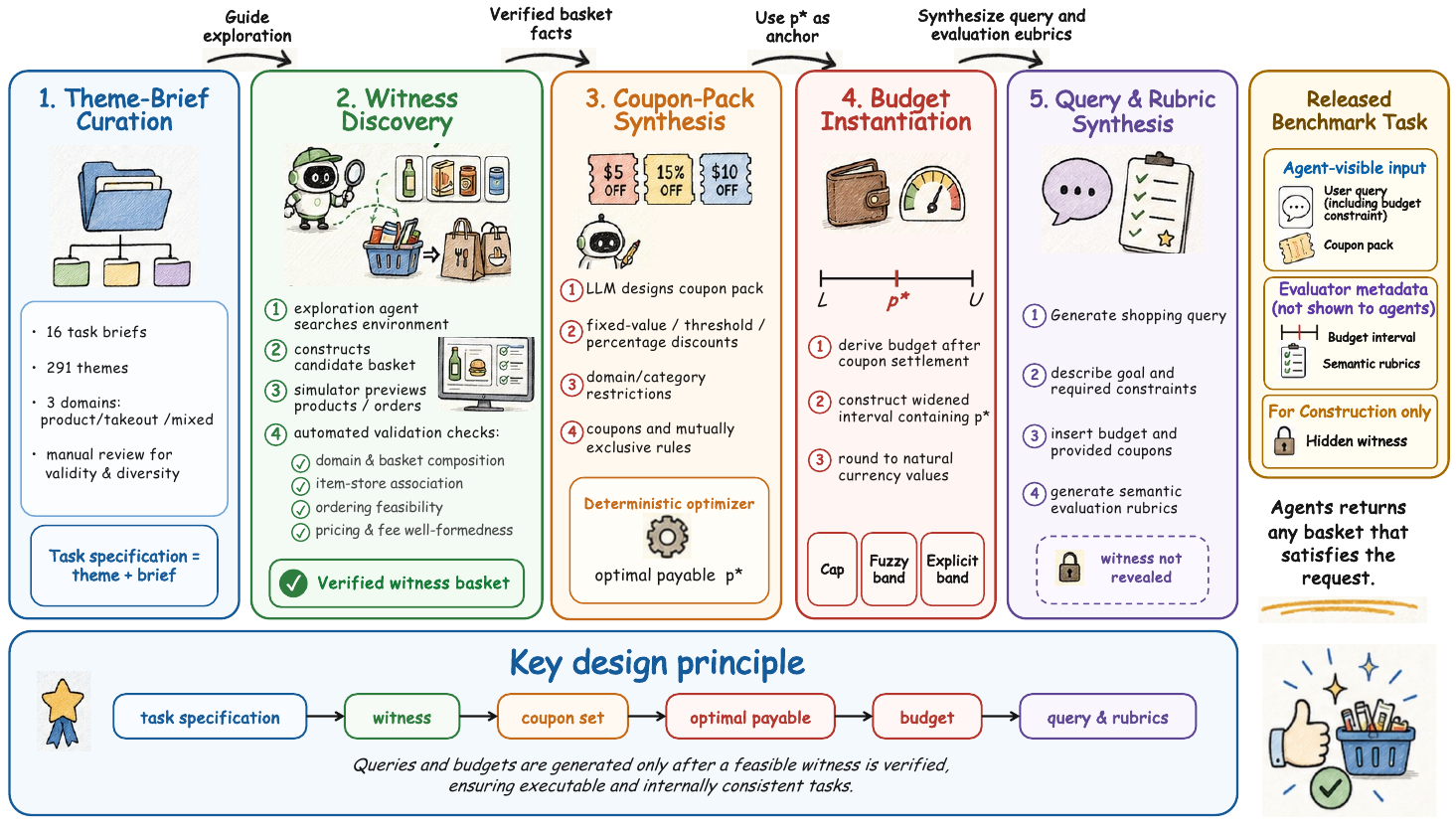}
\caption{Construction pipeline of \textsc{ComboShoppingBench}. From a theme--brief specification, the pipeline discovers a feasible hidden witness, then synthesizes coupons, a budget, and a query with semantic rubrics. This ordering ensures executability while allowing agents to return any valid basket.}
\label{fig:overview}
\end{figure*}

We first formulate the combo-shopping task, then describe task generation from hidden witness baskets and our framework for evaluating proposed baskets. Figure~\ref{fig:overview} summarizes the construction pipeline.

\subsection{Task Formulation}

We model the shopping agent as a tool-augmented policy $\pi_{\theta}$ that interacts with the commerce-and-takeout environment $\mathcal{E}$ to iteratively construct a shopping solution. At interaction step $t$, the agent conditions on the interaction history $h_t=(q,\mathcal{C},(a_i,o_i)_{i=1}^{t-1})$ and samples an action $a_t\sim\pi_{\theta}(\cdot\mid h_t)$, where each action corresponds to either a tool call (e.g., product search) or a terminal response. Executing a tool returns an observation $o_t$ from $\mathcal{E}$, which provides partial information about the underlying environment. After $T$ interaction steps, the resulting interaction trajectory $\tau_T=((a_t,o_t))_{t=1}^{T}$ produces the final output
\[
\hat{y}=(\hat{B},\hat{U},\hat{r}),
\]
where $\hat{B}$ denotes the recommended basket, $\hat{U}\subseteq\mathcal{C}$ is the selected coupon set, and $\hat{r}$ is the final response that presents and justifies the recommendation.


\subsection{Benchmark Construction}

\textbf{Theme--brief curation.}
To cover diverse and realistic combo-shopping scenarios, we first define 16
procurement briefs. A brief is a short design guideline for creating a
shopping request: it specifies the shopping domain, the intended roles and
approximate number of items, and the relationships that should hold among
them. For example, one brief asks for a core product together with accessories
whose interfaces or sizes depend on that product, while another asks for a
group meal containing complementary dishes and drinks. We then create 291
distinct themes, each describing a concrete shopping goal, such as
setting up a new phone or arranging a group meal, and pair each theme with a
suitable brief. We manually review and refine all theme--brief pairs. The final
collection contains 151 product-only, 90 takeout or instant-retail, and 50
mixed-domain tasks. Appendix~\ref{app:benchmark-details} reports the complete
brief coverage and task statistics.


\textbf{Solution-first construction.}
Given a task specification (i.e., a theme--brief pair), we first construct and validate a feasible basket as a witness. We then compute its costs, synthesize a coupon pack, determine the optimal legal coupon combination, and derive a feasible budget from the resulting payable amount. The query and semantic rubrics are generated only after these steps. The procedure is:
\[
\begin{aligned}
\text{task specification}
&\rightarrow \text{witness}
\rightarrow \text{coupon set} \\
&\rightarrow \text{optimal payable}
\rightarrow \text{budget} \\
&\rightarrow \text{query and rubrics}.
\end{aligned}
\]
This ordering ensures that every task is grounded in an executable scenario with verifiable ordering and payment.

\textbf{Witness discovery.}
Starting from a task specification, an exploration agent searches the environment and assembles a candidate basket containing commerce products, takeout or instant-retail orders, or a combination of both. The candidate basket then undergoes automated validation to determine whether it matches the specified shopping domain and basket composition, contains valid item--store associations, satisfies all ordering requirements, and includes well-formed price and fee information. When validation fails, the system returns concise diagnostic feedback to guide further exploration.

A validated basket serves as a witness, which provides evidence that the task is feasible. It is retained solely for task construction and auditing, is never revealed during evaluation, and is not treated as a reference solution. An evaluated agent may produce any basket that satisfies the generated request. Appendix F provides further evidence that successful agents need not reproduce the witness.

\textbf{Coupon-pack synthesis.}
For each witness basket, we deterministically compute merchandise and delivery costs. An LLM-based designer then creates a scenario-conditioned pack of five or six coupons. The pack includes threshold-reduction, direct-reduction, and capped percentage-discount coupons. Coupon scopes, thresholds, caps, and mutual-exclusion rules are varied to make the best combination non-obvious. The designer also includes plausible but suboptimal decoy coupons, turning coupon selection into a small combinatorial optimization problem. A deterministic optimizer identifies the optimal legal coupon combination for the witness. We denote the resulting final payable amount, including delivery fees, by $p^*$, which is used to determine the budget.

\textbf{Budget instantiation.}

We construct a moderately widened interval around $p^*$ and round its boundaries to natural currency values. The budget is expressed as a cap, a fuzzy target such as ``around $N$ CNY,'' or an explicit band. A fuzzy or explicit band specifies a target spending tier, whereas a cap imposes only an upper bound. All three modes are evaluated on the final payable after coupon settlement, including delivery fees.
Appendix~\ref{app:construction-details} describes the interval construction, rounding, and mode assignment.

\textbf{Query and rubric synthesis.}
Finally, an agent synthesizes a natural-language request from the task specification and the verified witness. The request states the shopping objective and constraints, incorporates the derived budget, and asks the evaluated agent to optimize coupon use. We then generate task-specific semantic rubrics that capture the key requirements used to assess whether a proposed basket fulfills the user's intent. Because the request is derived from a verified witness, each task has at least one feasible solution without requiring evaluated agents to reproduce the witness.

\subsection{Hybrid Evaluation}
We evaluate each final response along four complementary dimensions: semantic satisfaction measures alignment with the user's shopping intent; rule-based validation checks deterministic task constraints; response quality assesses presentation and usefulness; and claim faithfulness verifies factual statements against environment-recomputed results.

\textbf{Semantic satisfaction.}
A semantic judge evaluates the proposed basket against the task-specific rubrics synthesized for each query during benchmark construction. These rubrics capture the query's key semantic requirements, allowing this dimension to measure whether the selected items fulfill the user's intended objective while excluding deterministic ordering constraints and the correctness of explanatory claims.

\textbf{Rule-based validation.}
We extract the selected items, quantities, orders, and coupons from the final response and reconstruct the corresponding basket. A deterministic validation program then verifies order feasibility, coupon eligibility and optimality, and budget compliance using the simulator.

\textbf{Response quality.}
An LLM evaluator assesses whether the response clearly communicates the purchase plan, explains how the selected items support the user's objective, and concisely summarizes coupon and budget outcomes.

\textbf{Claim faithfulness.}
A claim-faithfulness judge evaluates the completeness and numerical accuracy of the monetary information reported in the final response. SKU-level prices and subtotals, realized coupon discounts, and final payable amounts are compared with values recomputed by deterministic code from the proposed basket. The response is considered faithful only when all required values are explicitly reported, correctly associated with the corresponding products or orders, and numerically accurate.

The prompts of the LLM evaluators described above are provided in Appendix~\ref{app:prompts}.

\section{Experiments}
\label{sec:experiments}

\begin{table*}[t!]
\caption{Pass rates (\%) on the 291-task \textsc{ComboShoppingBench}. Rule-based and Overall success are the per-task intersections defined in Section~\ref{sec:experiments}. ``Think'' and ``No-think'' denote inference configurations; the best results are bolded.}
\label{tab:main-results}
\centering
\small
\setlength{\tabcolsep}{4.2pt}
\renewcommand{\arraystretch}{1.04}
\resizebox{\textwidth}{!}{%
\begin{tabular}{@{}lccccccccc@{}}
\toprule
& \multicolumn{3}{c}{LLM-judged dimensions}
& \multicolumn{5}{c}{Rule-based validation} & \\
\cmidrule(lr){2-4}\cmidrule(lr){5-9}
Agent configuration
& Semantic
& \shortstack{Response\\quality}
& \shortstack{Claim\\faithfulness}
& \shortstack{Coupon-ID\\validity}
& \shortstack{Coupon\\legality}
& \shortstack{Coupon\\optimality}
& \shortstack{Budget\\compliance}
& Rule-based
& \shortstack{Overall\\success} \\
\midrule
GPT-5.5 (Think)                  & \textbf{83.8} & 92.4 & 90.4 & 99.3 & 96.6 & \textbf{94.2} & 86.3 & \textbf{83.8} & \textbf{61.2} \\
GPT-5.5 (No-think)               & 64.9 & 93.1 & 70.8 & 94.8 & 88.3 & 39.2 & 61.9 & 28.5 & 14.4 \\
\addlinespace[1.5pt]
GLM-5.2 (Think)                  & 75.3 & 83.2 & 90.4 & 96.6 & 95.5 & 91.8 & 83.5 & 80.8 & 52.9 \\
GLM-5.2 (No-think)               & 64.3 & 90.7 & 78.0 & 97.3 & 93.5 & 68.4 & 78.4 & 57.0 & 30.2 \\
\addlinespace[1.5pt]
Gemini-3.1-Pro (Think)           & 69.8 & 93.8 & 88.0 & 99.7 & 98.6 & 93.8 & \textbf{87.3} & 82.8 & 50.2 \\
Gemini-3.1-Pro (No-think)        & 60.1 & 88.7 & 78.7 & 98.6 & 95.5 & 79.4 & 82.8 & 70.4 & 35.7 \\
\addlinespace[1.5pt]
Claude-Opus-4.6 (Think)          & 77.7 & 92.1 & 94.5 & \textbf{100.0} & \textbf{99.7} & 93.1 & 80.4 & 73.9 & 49.8 \\
Claude-Opus-4.6 (No-think)       & 76.6 & 19.6 & \textbf{94.8} & 99.7 & 98.3 & 91.8 & 82.5 & 75.9 & 11.3 \\
\addlinespace[1.5pt]
Claude-Opus-4.8 (Think)          & 71.1 & \textbf{96.6} & 94.5 & \textbf{100.0} & 98.6 & 91.1 & 80.1 & 73.5 & 49.5 \\
Claude-Opus-4.8 (No-think)       & 63.6 & 70.1 & 78.7 & 98.6 & 95.9 & 76.6 & 73.5 & 57.0 & 21.3 \\
\addlinespace[1.5pt]
Kimi-K2.6 (Think)                & 72.5 & 92.8 & 78.4 & 96.9 & 93.8 & 77.0 & 80.8 & 67.4 & 44.0 \\
Kimi-K2.6 (No-think)             & 57.7 & 75.6 & 57.0 & 83.5 & 78.4 & 43.6 & 65.6 & 35.7 & 14.1 \\
\addlinespace[1.5pt]
DeepSeek-V4-Pro (Think)          & 69.1 & 89.3 & 80.1 & 96.9 & 93.5 & 77.3 & 78.7 & 63.6 & 34.4 \\
DeepSeek-V4-Pro (No-think)       & 67.4 & 14.1 & 68.4 & 93.1 & 88.7 & 61.9 & 72.5 & 48.8 & 4.1 \\
\addlinespace[1.5pt]
Claude-Sonnet-4.6 (Think)        & 70.1 & 59.8 & 83.2 & 99.7 & 96.2 & 83.8 & 74.9 & 65.6 & 27.8 \\
Claude-Sonnet-4.6 (No-think)     & 63.6 & 68.4 & 77.0 & 99.3 & 96.6 & 80.8 & 78.0 & 65.3 & 23.4 \\
\addlinespace[1.5pt]
MiniMax-M3 (Think)               & 71.8 & 72.5 & 68.0 & 88.0 & 80.8 & 67.4 & 66.3 & 54.0 & 25.4 \\
MiniMax-M3 (No-think)            & 63.9 & 44.3 & 52.9 & 93.8 & 78.4 & 40.9 & 58.8 & 31.3 & 6.2 \\
\addlinespace[1.5pt]
Doubao-Seed-2.0-Pro (Think)      & 56.7 & 81.8 & 69.4 & 96.2 & 85.2 & 56.0 & 72.9 & 46.4 & 17.5 \\
Doubao-Seed-2.0-Pro (No-think)   & 55.3 & 88.7 & 45.0 & 88.7 & 66.7 & 15.5 & 53.6 & 9.6 & 3.1 \\
\addlinespace[1.5pt]
Qwen3.6-27B (Think)              & 60.5 & 78.4 & 61.5 & 95.9 & 87.3 & 27.1 & 63.2 & 20.3 & 5.2 \\
Qwen3.6-27B (No-think)           & 62.5 & 51.2 & 81.1 & 98.3 & 95.2 & 64.3 & 74.6 & 49.1 & 12.4 \\
\bottomrule
\end{tabular}
}
\end{table*}

\subsection{Experimental Setup}

\textbf{Benchmark.}
The evaluated \textsc{ComboShoppingBench} contains 291 tasks: 151 product-only tasks, 90 takeout or instant-retail tasks, and 50 mixed-domain tasks. Each task provides a natural-language shopping request, five or six coupons, and a budget expressed as an upper limit, an approximate target, or an explicit range. Appendix~\ref{app:benchmark-details} reports the full benchmark composition.


\textbf{Agents and inference configurations.}
We evaluate 11 agents, each under provider-supported Think and No-think configurations, yielding 22 agent configurations. Every agent configuration receives the same tasks, coupons, tool interface, retrieval setup, output limit, and maximum number of turns. Appendices~\ref{app:environment} and~\ref{app:inference-details} provide the complete configurations.

\textbf{Construction and evaluation models.}
All benchmark-construction roles use GPT-5.5 with role-specific prompts. Final responses are evaluated using deterministic identifier validation and independent Gemini-3.1-Pro-Preview judgments of semantic satisfaction, response quality, and claim faithfulness. Full configurations and evaluator inputs are provided in Appendix~\ref{app:inference-details}.

\textbf{Metrics.}
Let $S$, $V$, $Q$, and $F$ denote semantic satisfaction, rule-based validation, response quality, and claim faithfulness, respectively. $S$ requires every query-specific semantic criterion to pass. $V$ requires valid coupon identifiers, a legal and basket-optimal coupon combination, and a final payable within the structured budget interval. $Q$ requires all five presentation criteria to pass, and $F$ requires all four settlement-disclosure criteria to pass. We report each dimension separately and define Overall Success as the intersection of $S$, $V$, $Q$, and $F$.
\begin{equation}
\text{Overall Success}=S\land V\land Q\land F.
\label{eq:experimental-metrics}
\end{equation}

\subsection{End-to-End Performance}

\begin{wrapfigure}{r}{0.58\textwidth}
\centering
\vspace{-4mm}
\includegraphics[width=\linewidth]{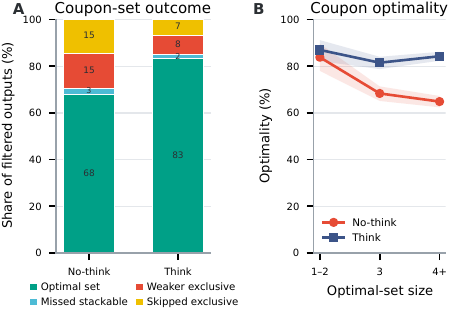}
\caption{Coupon selection analysis. (A) Coupon-set outcomes across the 11 No-think and 11 Think settings. (B) Coupon optimality by optimal-set size. Bands show 95\% task-cluster bootstrap intervals.}
\label{fig:coupon-analysis}
\end{wrapfigure}

As shown in Table~\ref{tab:main-results}, high pass rates on individual dimensions do not imply reliable end-to-end performance. GPT-5.5 (Think) achieves the best Overall success but still fails nearly 40\% of tasks despite exceeding 83\% on both Semantic and Rule-based validation and 90\% on response-related dimensions. The gap between marginal and joint pass rates shows that agents must maintain correctness across the entire pipeline. Error profiles also differ: Qwen3.6-27B (Think) usually produces valid coupon IDs but often selects suboptimal coupons, whereas Claude-Opus-4.6 (No-think) passes most rule-based checks but frequently fails in response quality. These differences motivate joint evaluation across all dimensions.

\subsection{Failure Analysis}
\label{sec:transactional-analysis}



\textbf{Agents struggle to choose mutually exclusive coupons.}
The main bottleneck in coupon reasoning lies in selecting the best coupon among mutually exclusive alternatives. As shown in Figure~\ref{fig:coupon-analysis}A, agents rarely miss stackable coupons; the performance gap mainly arises from decisions over exclusive coupons. Think reduces the rates of selecting a suboptimal exclusive coupon and skipping an exclusive group from 15\% each to 8\% and 7\%, respectively, increasing the optimal-set rate from 68\% to 83\%. Figure~\ref{fig:coupon-analysis}B further shows that coupon optimality under No-think declines markedly as the optimal-set size increases, whereas Think remains more stable. These results suggest that Think improves global planning over multiple coupons, while optimal selection within exclusive groups remains the central challenge.


\begin{wrapfigure}{r}{0.58\textwidth}
\centering
\vspace{-4.3mm}
\includegraphics[width=0.95\linewidth]{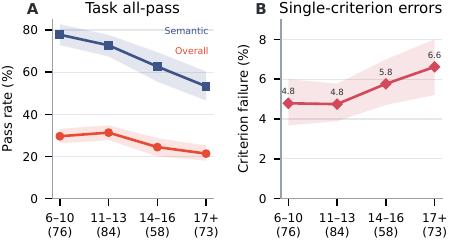}
\caption{Performance by semantic-criterion count: (A) task-level pass rates and (B) criterion-level failure rates.}
\label{fig:compositional-complexity}

\end{wrapfigure}

\textbf{The difficulty of compositional shopping stems primarily from constraint accumulation.} As the number of semantic criteria increases, the task-level Semantic all-pass rate drops substantially, while the criterion-level failure rate rises only from about 4.8\% to 6.6\% (Figure~\ref{fig:compositional-complexity}). This pattern suggests that models usually satisfy most requirements but are increasingly likely to miss a few constraints. Since all-pass requires every criterion to be satisfied, these errors accumulate as the number of constraints grows, producing task-level failures. The core challenge is therefore satisfying all requirements jointly.
\begin{figure}[H]
\centering
\includegraphics[width=\columnwidth]{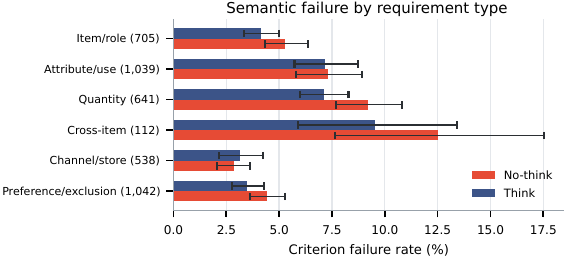}
\caption{Criterion failure rates by semantic requirement type. Parentheses give criterion counts; error bars show 95\% task-bootstrap intervals.}
\label{fig:semantic-taxonomy}
\end{figure}

\textbf{Cross-item requirements have the highest semantic failure rate.}
Semantic failures primarily arise from cross-item reasoning rather than single-item matching. We group task-specific semantic criteria by requirement type and report the criterion-level failure rate for each category. As shown in Figure~\ref{fig:semantic-taxonomy}, cross-item relations have the highest failure rate, followed by quantity requirements. The former require reasoning about compatibility and joint goal satisfaction, while the latter require accurate tracking of quantities and set size. In contrast, channel and store requirements are the easiest to satisfy. Thus, the main bottleneck is not finding relevant products but coordinating multiple products and their constraints. Think reduces failures in cross-item relations, quantity requirements, and item/role requirements, yet cross-item relations remain the most challenging. Models can often identify suitable individual products but struggle to ensure that the full set jointly satisfies the request.


\subsection{Does the Thinking Configuration Help?}
\label{sec:thinking-analysis}
\begin{wrapfigure}{r}{0.58\textwidth}
\centering
\includegraphics[width=\linewidth]{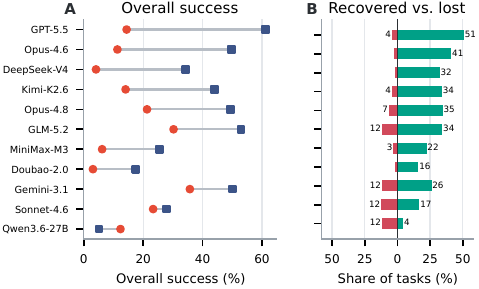}
\caption{Paired comparison of Think and No-think configurations. (A) Overall success. (B) Tasks lost or recovered under Think. Agents are ordered by the net change in Overall success.}
\label{fig:thinking-transitions}
\end{wrapfigure}

\textbf{Think improves aggregate performance for most agents, but its effect is not uniformly positive across tasks.} As shown in Figure~\ref{fig:thinking-transitions}, the overall gain comes from recovered tasks outnumbering lost tasks. Think changes the solution trajectory: it resolves some tasks that fail under No-think while causing a subset of previously successful tasks to fail. This pattern reflects the multi-step nature of compositional shopping. Think can improve planning, constraint checking, and coordination across intermediate decisions, but it may also alter search, product selection, coupon comparison, and tool-use behavior. Qwen3.6-27B illustrates this failure mode: it loses more tasks than it recovers. We find that its average calculator usage drops from 4.21 to 1.19 calls under Think. The accompanying increase in calculation errors suggests that the model may over-rely on internal reasoning at the expense of necessary exact computation.

\subsection{Are the LLM-Based Evaluators Reliable?}
\label{sec:evaluator-validity}
\begin{table}[t]
\caption{Rubric-level agreement (\%) between each LLM evaluator and the
expert-adjudicated human reference. $N$ denotes the number of human-reference
decisions.}
\label{tab:judge-human-agreement}
\centering
\footnotesize
\setlength{\tabcolsep}{5pt}
\renewcommand{\arraystretch}{1.06}
\begin{tabular}{@{}llc@{}}
\toprule
Dimension ($N$) & Evaluator & Agreement \\
\midrule
Semantic (436) & Gemini-3.1-Pro & 98.17 \\
& GPT-5.5 & 97.71 \\
& Kimi-K2.6 & 98.39 \\
\addlinespace[2pt]
Response Quality (150) & Gemini-3.1-Pro & 94.67 \\
& GPT-5.5 & 95.33 \\
& Kimi-K2.6 & 91.33 \\
\addlinespace[2pt]
Claim Faithfulness (114) & Gemini-3.1-Pro & 99.12 \\
& GPT-5.5 & 99.12 \\
& Kimi-K2.6 & 99.12 \\
\addlinespace[2pt]
Overall (700) & Gemini-3.1-Pro & 97.57 \\
& GPT-5.5 & 97.43 \\
& Kimi-K2.6 & 97.00 \\
\bottomrule
\end{tabular}
\end{table}

To assess the reliability of our LLM-based evaluators, we construct a human-annotated reference set from 30 outputs generated by Qwen3.6-27B (No-think). The samples are stratified to cover successful cases and diverse failure modes. Two annotators independently label every rubric instance while blinded to both the evaluated model and the LLM evaluators' predictions. An expert adjudicates all disagreements, yielding 700 reference decisions: 436 for Semantic, 150 for Response Quality, and 114 for Claim Faithfulness. We then use Gemini-3.1-Pro, GPT-5.5, and Kimi-K2.6 as independent LLM evaluators. Each evaluator judges all 700 rubric instances from the same 30 samples. We measure reliability by rubric-level agreement, defined as the percentage of evaluator decisions that exactly match the expert-adjudicated human reference. The complete annotation protocol and additional agreement analyses are provided in Appendix~\ref{app:human-eval}.

\textbf{LLM evaluators closely match expert-adjudicated human judgments.}
As shown in Table~\ref{tab:judge-human-agreement}, all three evaluators achieve at least 97\% overall agreement with the human reference. The agreement is consistently high for Semantic (97.71--98.39\%) and reaches 99.12\% for Claim Faithfulness across all evaluators. Response Quality is more subjective and exhibits greater variation, but agreement remains above 91\%.

The evaluators also achieve 97.14--98.14\% pairwise agreement, with detailed results reported in Appendix~\ref{app:human-eval}.

\section{Conclusion}
We introduced \textsc{ComboShoppingBench}, a benchmark for open-ended, budget-constrained basket shopping with coupons. We proposed a solution-first pipeline to construct feasible tasks from validated hidden witness baskets and a hybrid evaluation framework that combines LLM-based evaluators with deterministic validation. The LLM evaluators achieve 97.00--97.57\% agreement with expert-adjudicated human judgements. Across 291 tasks and 22 configurations of 11 agents, the strongest configuration achieves only 61.2\% Overall success. Our analysis identify cross-item constraint accumulation and optimization over mutually exclusive coupons as key bottlenecks and show that Think configurations generally help but are not uniformly beneficial. 

\bibliographystyle{unsrtnat}
\bibliography{references}

\FloatBarrier
\clearpage

\begin{center}
  {\titlefont\sffamily\bfseries
  ComboShoppingBench: Evaluating LLM Agents for Budget-Constrained Basket Shopping with Coupons Supplementary Material\par}
  \vspace{4mm}
  {\authorlist\par}
  \vspace{2mm}
  {\affiliationlist\par}
\end{center}

\renewcommand{\contentsname}{Supplementary Contents}
\addtocontents{toc}{\protect\setcounter{tocdepth}{2}}
\setcounter{tocdepth}{2}
\tableofcontents
\vspace{0.5em}

\setcounter{secnumdepth}{2}
\appendix
\numberwithin{figure}{section}
\numberwithin{table}{section}
\numberwithin{equation}{section}

\section{Benchmark Composition and Task Statistics}
\label{app:benchmark-details}

\begin{figure*}[!t]
\centering
\includegraphics[width=0.96\textwidth]{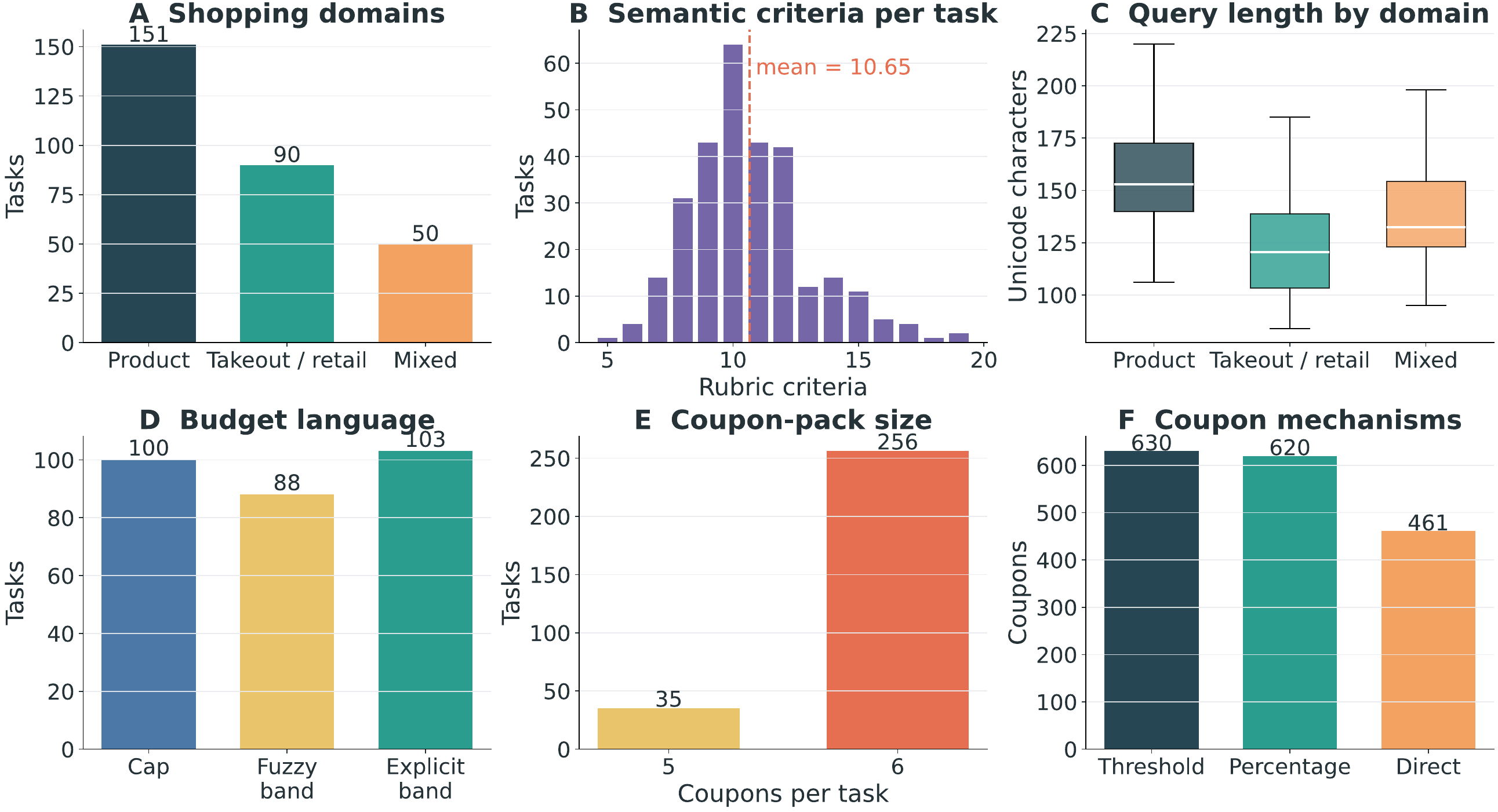}
\caption{Overview of the 291 benchmark tasks, including shopping-domain
coverage, the number of semantic criteria per task, request length, budget
wording, coupon-pack size, and coupon mechanisms.}
\label{fig:benchmark-overview}
\end{figure*}

We summarize the 291 benchmark tasks used in all experiments. The
statistics below are computed from the shopping requests, the semantic
criteria used to judge them, the coupon packs, and the stated budgets. They do
not use agent responses or agent performance. Figure~\ref{fig:benchmark-overview}
provides an overview of the main task characteristics, and
Table~\ref{tab:benchmark-domain-stats} gives the exact composition by shopping
domain.

\begin{table*}[!t]
\caption{Composition of the 291-task benchmark. Request length counts written
characters, including letters, numbers, and punctuation. The final three
columns report averages within each shopping domain.}
\label{tab:benchmark-domain-stats}
\centering
\small
\setlength{\tabcolsep}{7pt}
\renewcommand{\arraystretch}{1.14}
\begin{tabular}{@{}lrrrrr@{}}
\toprule
Domain & Tasks & Share & Request length & Semantic criteria & Coupons \\
\midrule
Product only & 151 & 51.9\% & 159.1 & 10.97 & 5.88 \\
Takeout / instant retail & 90 & 30.9\% & 124.1 & 9.57 & 5.87 \\
Mixed domain & 50 & 17.2\% & 139.9 & 11.66 & 5.90 \\
\midrule
Overall & 291 & 100.0\% & 145.0 & 10.65 & 5.88 \\
\bottomrule
\end{tabular}
\end{table*}

\subsection{Shopping Goals and Procurement Briefs}

Each task begins with a concrete shopping goal, such as equipping a new phone,
preparing a meal for one person, or arranging supplies for an event. The 291
tasks use 291 different goals. To make these goals cover different kinds of
multi-item shopping, we organize them with 16 \emph{procurement briefs}. A
procurement brief is a short instruction used when creating a task. It states
what kind of request to write: which shopping channel to use, which item roles
the request should contain, and how the items should work together. For
example, a brief may call for one main product and several accessories that
must fit it, or for a group meal containing main dishes, sides, staples, and
drinks. Evaluated agents receive the resulting shopping request, not the
brief itself.

Table~\ref{tab:procurement-briefs} lists the coverage of all 16 briefs. The item
counts in the descriptions express the intended shape of the requests. They
do not require every valid agent solution to contain exactly that many items.

\begin{table*}[!t]
\caption{Coverage of the 16 procurement briefs used to create the benchmark
tasks. A brief describes the intended shopping need and the relationships
among the requested items.}
\label{tab:procurement-briefs}
\centering
\footnotesize
\setlength{\tabcolsep}{4pt}
\renewcommand{\arraystretch}{1.08}
\resizebox{\textwidth}{!}{%
\begin{tabular}{@{}>{\raggedright\arraybackslash}p{0.12\textwidth}>{\raggedright\arraybackslash}p{0.22\textwidth}r>{\raggedright\arraybackslash}p{0.57\textwidth}@{}}
\toprule
Domain & Procurement brief & Tasks & Intended shopping need \\
\midrule
Product
& Core product and required accessories
& 42
& One main product with two or three necessary accessories whose model, interface, size, or installation method depends on the main product. \\

Product
& Complementary product set
& 59
& Four or five products that complete one practical task, including at least one compatibility, connection, installation, containment, or workflow relationship. \\

Product
& Large complementary set
& 15
& Six to eight products forming a complete setup, with at least two explicit relationships among components or task steps. \\

Product
& Books and study materials
& 8
& Three to five printed resources aligned by edition, grade, subject, practice format, or stage of study. \\

Product
& Consumable replenishment
& 15
& Four to six frequently used supplies, including items that fit the same device, model, or primary consumable. \\

Product
& Professional tools and materials
& 12
& Four to six tools, materials, or components covering the steps of one technical task and matching in specification or material. \\
\midrule
Takeout
& Individual meal
& 25
& Three or four dishes or drinks forming a complete and appropriately sized meal for one person. \\

Takeout
& Group meal
& 17
& Five to seven shareable items covering main dishes, sides, staples, and drinks for three to five people. \\

Instant retail
& Immediate non-food supplies
& 36
& Two to four quickly delivered products that solve one concrete problem through paired use, processing steps, or complementary functions. \\

Instant retail
& Gift set
& 4
& Two to four products combining a main gift with suitable wrapping, presentation, or quantity for the recipients. \\

Instant retail
& Drinks and party food
& 3
& Three to five drinks and foods coordinated by number of guests, serving quantity, or pairing. \\

Instant retail
& Pet or baby supplies
& 5
& Two to four urgent supplies supporting one feeding, cleaning, care, or settling need. \\
\midrule
Mixed
& Direct cross-channel pairing
& 15
& One to three immediately delivered products plus one or two conventional-commerce products used directly to prepare, handle, or consume them. \\

Mixed
& Two-channel shopping task
& 10
& Two or three conventional-commerce products and two or three immediately delivered products that jointly complete one task. \\

Mixed
& Emergency care
& 10
& One or two tools or supplies from conventional commerce plus two to four immediate supplies addressing the same urgent problem. \\

Mixed
& Event preparation
& 15
& Two or three decorations or tools from conventional commerce plus two to four immediate products linked by serving, display, preparation, or guest-count requirements. \\
\bottomrule
\end{tabular}
}
\end{table*}

The brief distribution is designed for coverage rather than equal class sizes.
The largest brief contains 59 tasks, whereas the smallest contains three.
Accordingly, the overall benchmark score reflects the full task mixture; a
comparison among individual briefs would require care because several briefs
contain only a small number of tasks.

\subsection{Shopping Requests and Semantic Criteria}

The shopping requests contain 145.0 Unicode characters on average, with a
standard deviation of 31.3 and a range from 84 to 282. Product-only requests
are longer on average than takeout or instant-retail requests, while
mixed-domain requests fall between them. Character count is reported only as
a description of request length and is not treated as a measure of difficulty.

Across the benchmark, the semantic judge uses 3,100 task-specific criteria.
Each task has 10.65 criteria on average, with a standard deviation of 2.41, a
median of 10, an interquartile range of 9--12, and a range of 5--19. These
criteria turn each shopping request into concrete checks: whether all required
item roles are present; whether brands, models, specifications, quantities,
serving sizes, or recipient counts are respected; whether accessories fit the
main product; whether foods and drinks form an appropriate meal; and whether
items purchased through different channels jointly serve the stated goal.
The number of criteria describes how many aspects of a request are checked; it
does not by itself establish that one task is harder than another.

\subsection{Coupon Packs and Budget Expressions}

Every task supplies a coupon pack and states the budget directly in the
shopping request. The benchmark contains 1,711 coupons in total. Thirty-five
tasks provide five coupons and 256 provide six. Coupon decisions vary in three
ways: how the discount is calculated, which products it applies to, and
whether it can be combined with other coupons. Table~\ref{tab:coupon-budget-stats}
reports the resulting counts.

\begin{table*}[!t]
\caption{Coupon and budget composition. Shares for coupon mechanism, scope,
and mutual exclusion are calculated over 1,711 coupons; shares for coupon-pack
size and budget wording are calculated over 291 tasks.}
\label{tab:coupon-budget-stats}
\centering
\small
\setlength{\tabcolsep}{8pt}
\renewcommand{\arraystretch}{1.12}
\begin{tabular}{@{}llrr@{}}
\toprule
Dimension & Category & Count & Share \\
\midrule
Coupons per task & Five & 35 tasks & 12.03\% \\
& Six & 256 tasks & 87.97\% \\
\midrule
Discount mechanism & Threshold reduction & 630 coupons & 36.82\% \\
& Percentage discount & 620 coupons & 36.24\% \\
& Direct reduction & 461 coupons & 26.94\% \\
\midrule
Eligible products & All products & 631 coupons & 36.88\% \\
& One shopping domain & 302 coupons & 17.65\% \\
& One product category & 778 coupons & 45.47\% \\
\midrule
Combination rule & Belongs to an exclusion group & 1,201 coupons & 70.19\% \\
& No exclusion group & 510 coupons & 29.81\% \\
\midrule
Budget wording & Upper limit & 100 tasks & 34.36\% \\
& Approximate target & 88 tasks & 30.24\% \\
& Explicit range & 103 tasks & 35.40\% \\
\bottomrule
\end{tabular}
\end{table*}

For a threshold-reduction coupon, the eligible merchandise must reach a stated
minimum before a fixed amount is deducted. A percentage coupon reduces the
eligible subtotal, sometimes subject to a maximum discount, whereas a direct
reduction deducts a fixed amount without a spending threshold. Coupons assigned
to the same exclusion group cannot be used together.

An upper-limit request asks the agent to spend no more than a stated amount;
an approximate-target request asks it to stay around an amount; and an
explicit-range request gives both a lower and an upper bound. All three refer
to the final amount paid after coupon discounts and delivery fees. The median
target amounts are CNY~300 for product-only tasks, CNY~31.5 for takeout or
instant-retail tasks, and CNY~90 for mixed-domain tasks. These values mainly
reflect the different prices of the goods involved and should not be read as a
cross-domain measure of difficulty.

\section{E-Commerce and Takeout Environment}
\label{app:environment}

\subsection{Environment Snapshot}

\textsc{ComboShoppingBench} uses a fixed, offline, and read-only catalog snapshot spanning conventional commerce and takeout or instant retail. The environment provides stable product facts, store-level constraints, retrieval, and deterministic order-feasibility checks without accessing an online shopping service. These checks do not trigger payment, inventory changes, or fulfillment. Table~\ref{tab:environment-scale} summarizes the catalog objects used throughout task construction and evaluation.

\begin{table*}[!t]
\caption{Scale and roles of the frozen catalog snapshot used in the benchmark.}
\label{tab:environment-scale}
\centering
\footnotesize
\setlength{\tabcolsep}{3pt}
\renewcommand{\arraystretch}{1.16}
\begin{tabular}{@{}>{\raggedright\arraybackslash}p{0.22\textwidth}>{\raggedright\arraybackslash}p{0.13\textwidth}>{\raggedright\arraybackslash}p{0.39\textwidth}>{\raggedright\arraybackslash}p{0.20\textwidth}@{}}
\toprule
Catalog object & Scale & Decision-relevant information & Role in the benchmark \\
\midrule
Commerce products
& 4,306,132 SKUs
& Product identifiers, titles, brands, categories, prices, attributes, tags, and review summaries
& Commerce product retrieval and basket construction \\

Takeout and instant-retail stores
& 5,171 stores
& Store identifiers, business metadata, delivery fees, free-delivery thresholds, and minimum-order thresholds
& Store selection and store-level feasibility constraints \\

Takeout and instant-retail products
& 341,204 SKUs
& Store-linked product identifiers, names, prices, and category metadata
& Menu or local-retail retrieval and same-store order construction \\
\bottomrule
\end{tabular}
\end{table*}

Catalog facts are stored in a local relational database and paired with a dense vector index for semantic retrieval. The environment supports lexical, vector, and hybrid retrieval, but the benchmark fixes all evaluated agents to vector retrieval so that retrieval behavior does not vary with agent-selected search modes. Observations omit large raw records and retrieval-debugging fields while retaining the identifiers, attributes, prices, and store rules needed for selection and settlement reasoning.

\subsection{Agent Tool Interface}
\label{app:tool-interface}

All evaluated agents interact with the same four tools: three retrieval tools over the simulated commerce and takeout environment, and a restricted Python calculator for deterministic arithmetic. Table~\ref{tab:agent-tools} summarizes their roles, agent-visible inputs, returned information, and fixed retrieval budgets.

\begin{table*}[!t]
\caption{Tools available to evaluated agents. Retrieval limits are fixed across all agent configurations.}
\label{tab:agent-tools}
\centering
\footnotesize
\setlength{\tabcolsep}{2.5pt}
\renewcommand{\arraystretch}{1.16}
\begin{tabular}{@{}>{\raggedright\arraybackslash}p{0.19\textwidth}>{\raggedright\arraybackslash}p{0.16\textwidth}>{\raggedright\arraybackslash}p{0.17\textwidth}>{\raggedright\arraybackslash}p{0.28\textwidth}>{\raggedright\arraybackslash}p{0.14\textwidth}@{}}
\toprule
Tool & Purpose & Agent-visible input & Returned information & Fixed budget or restriction \\
\midrule
\texttt{product\_search}
& Search the commerce catalog
& Natural-language query list
& SKU identifiers, prices, and product metadata
& 8 queries; 8 products per query; 45 products total \\

\texttt{takeout\_search}
& Search across takeout and instant-retail stores
& Natural-language query list
& Store identifiers, delivery and minimum-order rules, and candidate SKUs
& 8 stores per query; 8 stores total; 5 SKUs per store \\

\shortstack[l]{\texttt{takeout\_search}\\\texttt{\_in\_store}}
& Search within a selected store
& Store identifier and one or more queries
& Store metadata and matched SKU identifiers and prices
& 6 queries; 8 SKUs per query \\

\texttt{python\_calculator}
& Perform exact arithmetic
& Short Python code
& Text explicitly emitted by \texttt{print()}
& Restricted built-ins; no imports, files, or network access \\
\bottomrule
\end{tabular}
\end{table*}

\paragraph{Search tools.}
\texttt{product\_search} retrieves ordinary e-commerce products. For takeout and instant-retail requests, \texttt{takeout\_search} first retrieves candidate stores together with store-level ordering constraints and a small set of relevant SKUs; after selecting a store, the agent can use \texttt{takeout\_search\_in\_store} to complete a same-store order. Search observations retain decision-relevant identifiers, prices, product attributes, and store settlement rules, while results from multiple queries are merged and deduplicated.

\paragraph{Calculator.}
The calculator executes short, pure-Python calculations in a restricted environment and returns only text produced by \texttt{print()}. The \texttt{python\_calculator} tool cannot access the product environment or replace the deterministic validator. Across all agents, retrieval budgets and schemas are identical, transient infrastructure failures receive at most one retry, and each serialized tool observation is capped at 80,000 characters. The tools provide candidate facts and arithmetic support; determining semantic suitability, coupon validity and optimality, and budget compliance remains the responsibility of the agent and the evaluation pipeline.

\subsection{Order-Feasibility Validation}

The \emph{Order feasibility} component reported in the main text is computed by two deterministic validators, one for commerce products and one for takeout or instant-retail orders. These validators are not exposed to evaluated agents. After extracting an agent's final selection, the evaluator invokes them internally to reconstruct orders from catalog prices and store rules. For a set $P$ of commerce products with snapshot price $p_i$ and quantity $q_i$, the merchandise subtotal is
\begin{equation}
S_{\mathrm{com}}=\sum_{i\in P}p_iq_i.
\label{eq:commerce-subtotal}
\end{equation}
The commerce feasibility check validates product identifiers and quantities and recomputes line totals and the subtotal; the environment does not model commerce shipping fees.

Takeout and instant-retail products may be purchased from multiple stores, but each store forms a separate order. For store $j$, let $O_j$ denote its selected products, $B_j$ its base delivery fee, $T_j$ its free-delivery threshold, and $M_j$ its minimum-order threshold. The takeout feasibility check computes
\begin{equation}
\begin{aligned}
S_j &= \sum_{i\in O_j}p_iq_i, \\
F_j &=
\begin{cases}
0, & S_j\ge T_j,\\
B_j, & S_j<T_j,
\end{cases} \\
E_j &= S_j+F_j, \qquad S_j\ge M_j.
\end{aligned}
\label{eq:takeout-settlement}
\end{equation}
An order is infeasible if its products do not belong to the same store, contain invalid identifiers, or do not meet $M_j$. A multi-store solution passes \emph{Order feasibility} only if every store-specific order is feasible; merchandise and delivery fees are then summed across orders.

Coupons and budgets are introduced by the benchmark's task-generation layer rather than by the catalog environment. Coupon discounts apply only to eligible merchandise amounts; delivery fees are added after discounting. Thus, the environment establishes product facts and order feasibility, while the benchmark layer defines the coupon and final-payable constraints used for task construction and evaluation.

\subsection{System Boundary and Limitations}

The environment is responsible for catalog facts, retrieval, SKU reconstruction, and deterministic order-feasibility validation. The task generator adds scenario-specific requests, coupons, budgets, semantic rubrics, and a hidden feasible witness; the evaluator extracts each agent's proposed basket and independently reruns order and settlement checks. Because the environment is based on a fixed snapshot, it does not model real-time inventory or price changes, personalization, address-dependent availability, distance, weather, dynamic delivery capacity, payment, or fulfillment. It therefore supports controlled evaluation of catalog search, basket construction, constraint satisfaction, and settlement reasoning, but does not measure online conversion or real-world delivery performance.

\section{Coupon Design and Settlement Rules}
\label{app:coupon-details}

Each task provides the evaluated agent with a coupon pack in addition to the
shopping request. Coupons are introduced by the benchmark rather than by the
underlying catalog environment. They are designed to test three related
decisions: whether a coupon applies to the selected products, whether it can
be combined with the other selected coupons, and whether the resulting legal
combination achieves the lowest payable amount for the proposed basket.

\paragraph{Coupon-pack design.}
For each hidden witness basket, an LLM-based designer constructs a
scenario-conditioned pack of five or six coupons. A pack combines threshold,
fixed, and percentage discounts and varies their thresholds, eligible
products, discount caps, and mutual-exclusion rules. It also contains
plausible but suboptimal coupons. For example, a coupon may advertise a large
percentage reduction but have a low discount cap, cover only a narrow product
category, or conflict with a more valuable coupon. These choices make the
best combination non-obvious: using every coupon or selecting the coupon with
the largest advertised reduction need not minimize the final payment.

\paragraph{Coupon types and scopes.}
Table~\ref{tab:coupon-definitions} defines the three discount mechanisms. Let
$S_c$ denote the original subtotal of the products covered by coupon $c$, and
let $\widetilde{S}_c$ denote their current amount after earlier coupons have
been applied. A coupon can cover the complete basket, one shopping domain
(commerce or takeout), one commerce category, or one broad takeout category.
Thus, products outside the stated scope neither contribute to a spending
threshold nor receive the corresponding discount.

\begin{table*}[!t]
\caption{Coupon types and their settlement semantics. A percentage value
$r$ denotes the fraction paid; for example, $r=0.85$ represents a 15\%
discount.}
\label{tab:coupon-definitions}
\centering
\small
\setlength{\tabcolsep}{6pt}
\renewcommand{\arraystretch}{1.15}
\begin{tabular}{@{}>{\raggedright\arraybackslash}p{0.18\textwidth}>{\raggedright\arraybackslash}p{0.31\textwidth}>{\raggedright\arraybackslash}p{0.42\textwidth}@{}}
\toprule
Type & Applicability & Realized discount \\
\midrule
Threshold discount
& The original eligible subtotal $S_c$ reaches the stated threshold.
& A fixed amount is deducted from the current eligible amount. \\

Fixed discount
& The coupon covers at least one selected product.
& A fixed amount is deducted without a spending threshold. \\

Percentage discount
& The coupon covers at least one selected product.
& The discount is $\widetilde{S}_c(1-r)$, optionally limited by a stated maximum. \\
\bottomrule
\end{tabular}
\end{table*}

\paragraph{Combination and settlement.}
Coupons are settled in two stages. The evaluator first applies fixed and
threshold discounts. A threshold is checked against the original subtotal
within that coupon's scope, while the realized reduction is deducted only
from the current amount of the covered products. It then applies percentage
discounts to the remaining current amounts within their respective scopes.
For a percentage coupon with payment ratio $r_c$ and maximum discount $K_c$,
the realized reduction is
\begin{equation}
D_c=\min\!\left(\widetilde{S}_c(1-r_c),K_c\right),
\label{eq:percentage-coupon}
\end{equation}
where the cap is omitted when the coupon has no maximum. When a coupon covers
only part of the basket, its discount is assigned to the covered products so
that later coupons operate on the updated amounts. Multiple coupons within
the same stage follow their order in the released coupon pack.

Coupons assigned to the same exclusion group cannot be used together;
coupons in different groups may be combined. Delivery fees are added after
all merchandise discounts and are not themselves discountable. Accordingly,
the settlement identity is
\begin{equation}
P(B,C)=S(B)-\sum_{c\in C}D_c+F(B),
\label{eq:coupon-settlement}
\end{equation}
where $B$ is the selected basket, $C$ is a legal coupon combination, $S(B)$
is its merchandise subtotal, and $F(B)$ is its total delivery fee.

\paragraph{Legality and basket-specific optimality.}
Coupon optimality is evaluated for the agent's own basket, not by matching
the coupon set used by the hidden witness. After reconstructing the proposed
basket, the deterministic evaluator checks coupon identifiers, scope and
threshold eligibility, and mutual-exclusion constraints. It then finds the
lowest payment among the legal combinations,
\begin{equation}
P_{\mathrm{opt}}(B)=
\min_{C\in\mathcal{C}_{\mathrm{legal}}(B)} P(B,C),
\label{eq:coupon-optimality}
\end{equation}
and compares the agent's selected combination with this value. This design
allows an agent to choose any semantically suitable basket while still making
coupon legality and optimality deterministically verifiable.

\paragraph{Illustrative settlement.}
Suppose the eligible merchandise subtotal is CNY~150. A threshold coupon
reduces CNY~20 when the subtotal reaches CNY~120, and a compatible percentage
coupon provides 10\% off with a CNY~20 cap. The threshold coupon is applied
first, reducing the merchandise amount to CNY~130. The percentage coupon then
reduces CNY~13, giving a post-coupon merchandise amount of CNY~117. With a
CNY~5 delivery fee, the final payable is CNY~122.

\section{Budget Construction and Operationalization}
  \label{app:construction-details}

  \paragraph{Settlement quantity.}
  Let $p^*$ denote the minimum final payable amount of the hidden witness
  basket under a legal and optimal coupon combination. The final payable
  includes delivery fees:
  \[
  \begin{aligned}
  p^*={}&\text{merchandise subtotal}\\
        &-\text{realized coupon discounts}\\
        &+\text{delivery fees}.
  \end{aligned}
  \]
  The witness is used only to ensure task feasibility and instantiate the
  budget; evaluated agents are not required to reproduce it.

  \paragraph{Adaptive monetary granularity.}
  To obtain natural currency expressions while accommodating the different
  price scales of commerce and takeout tasks, we define the rounding
  granularity
  \[
  s(p^*) =
  \begin{cases}
  1,  & p^* < 50,\\
  5,  & 50 \le p^* < 200,\\
  10, & 200 \le p^* < 1000,\\
  50, & p^* \ge 1000.
  \end{cases}
  \]
  We first construct an unrounded tolerance radius
  \[
  \delta(p^*)=\max(0.15p^*,5),
  \]
  and round the two endpoints outward:
  \[
  \begin{aligned}
  L & = \max\left\{1,\;
      s\left\lfloor\frac{p^*-\delta(p^*)}{s}\right\rfloor\right\},\\
  U & = s\left\lceil\frac{p^*+\delta(p^*)}{s}\right\rceil.
  \end{aligned}
  \]
  The displayed approximate target is
  \[
  N =
  s\,\operatorname{round}\left(\frac{p^*}{s}\right),
  \]
  where ties are resolved using round-to-even, following the implementation.
  Outward rounding ensures that the feasible witness remains inside
  $[L,U]$.
  
\paragraph{Tolerance calibration.}
To operationalize the natural-language expression “about \(N\) yuan,” we conducted a questionnaire study with 35 participants. Respondents were shown shopping requests with target amounts covering the price scales in our benchmark and were asked to specify the lowest and highest final payments they would still consider to be “about” the stated amount. The median acceptable deviations were 17\% below and 13\% above the target. Based on this population-level interval, we adopt a symmetric tolerance of 15\%. For low-price tasks, we impose a minimum radius of CNY 5 to avoid unrealistically narrow intervals caused by percentage scaling.

  \paragraph{Budget modes.}
  Each task is assigned one of three budget modes. Table~\ref{tab:budget-modes}
  summarizes the natural-language realization and its structured
  interpretation. Here, $P$ denotes the evaluator-recomputed final payable
  amount for the agent's own basket.

  \begin{table*}[!t]
  \caption{Construction and operational interpretation of the three budget
  modes. Monetary constraints are evaluated on the final payable after coupon
  discounts and delivery fees.}
  \label{tab:budget-modes}
  \centering
  \small
  \setlength{\tabcolsep}{6pt}
  \renewcommand{\arraystretch}{1.15}
  \begin{tabular}{@{}p{0.17\textwidth}p{0.31\textwidth}
  p{0.18\textwidth}p{0.25\textwidth}@{}}
  \toprule
  Mode & Query realization & Displayed values & Operationalized interval \\
  \midrule
  Upper limit
  & ``The final payable should not exceed $U$ CNY.''
  & Upper limit $U$
  & $P \le U$ \\

  Approximate target
  & ``The budget is around $N$ CNY.''
  & Rounded target $N$
  & Target-band adherence: $L \le P \le U$ \\

  Explicit range
  & ``The final payable should be between $L$ and $U$ CNY.''
  & Both $L$ and $U$
  & $L \le P \le U$ \\
  \bottomrule
  \end{tabular}
  \end{table*}

  \paragraph{Deterministic mode assignment.}
  Budget wording is assigned independently of agent outputs. For task
  identifier $i$, we compute
  \[
  z_i =
  \frac{
  \operatorname{uint64}
  \left(
  \operatorname{SHA256}(\textit{seed}\Vert\texttt{:}\Vert i)_{1:8}
  \right)
  }{2^{64}}.
  \]
  The mode is upper limit if $z_i<1/3$, approximate target if
  $1/3\le z_i<2/3$, and explicit range otherwise. Thus, each mode has equal
  assignment probability, while the fixed seed makes the assignment fully
  reproducible.

\paragraph{Interpretive scope.}
An approximate target denotes a target
spending tier, not merely a maximum. Its lower endpoint
prevents substantially cheaper solutions that imply a
different product tier, quantity, or level of completeness.
Coupon optimality is basket-specific: the agent must construct
a suitable basket within [L, U] and then minimize payment for
that basket. Semantic adequacy is assessed separately.

  \paragraph{Illustrative example.}
  For a witness with $p^*=298.14$, the construction uses $s=10$ and
  $\delta=44.72$, producing $L=250$, $N=300$, and $U=350$. Consequently, CNY 250 satisfies the target band for “around CNY 300,” whereas a final payment below CNY 250 undershoots the requested spending tier.

\section{Model-Role and Agent Inference Configurations}
\label{app:inference-details}
\subsection{Construction and Evaluation Models}

Table~\ref{tab:model-role-configurations} reports the language models used by
the benchmark-construction and evaluation pipelines. All four construction
roles use GPT-5.5, but each is invoked separately with a role-specific system
prompt and only the inputs required by that stage. For the main experimental
results, GPT-5.5 in No-think mode extracts the final selected basket and coupon
identifiers from the agent response. Gemini-3.1-Pro-Preview then independently
performs the three LLM-based evaluations. These evaluators are not ensembled:
each dimension is produced by one configured evaluator rather than a
multi-model vote.

\begin{table*}[!t]
\caption{Language-model configurations for benchmark construction and
evaluation. ``Default'' means that the corresponding field is not explicitly
overridden and the provider default is used. Maximum output is measured in
tokens per model invocation.}
\label{tab:model-role-configurations}
\centering
\footnotesize
\setlength{\tabcolsep}{2.5pt}
\renewcommand{\arraystretch}{1.13}
\begin{tabular}{@{}>{\raggedright\arraybackslash}p{0.105\textwidth}>{\raggedright\arraybackslash}p{0.13\textwidth}>{\raggedright\arraybackslash}p{0.14\textwidth}>{\raggedright\arraybackslash}p{0.13\textwidth}>{\raggedleft\arraybackslash}p{0.11\textwidth}>{\raggedright\arraybackslash}p{0.10\textwidth}>{\raggedright\arraybackslash}p{0.195\textwidth}@{}}
\toprule
Stage & Role & Model & Reasoning control & Max output per turn & Temperature & Procedure or pass rule \\
\midrule
Construction
& Witness explorer
& GPT-5.5
& Medium
& 3,000
& Default
& Up to 16 ReAct steps and three validation-guided basket attempts. \\

Construction
& Coupon designer
& GPT-5.5
& Medium
& 3,000
& Default
& Up to three synthesis attempts, with deterministic coupon-pack validation after each attempt. \\

Construction
& Query writer
& GPT-5.5
& Medium
& 3,000
& Default
& Generates the request from the verified witness, task specification, and structured budget. \\

Construction
& Semantic-rubric writer
& GPT-5.5
& Medium
& 3,000
& Default
& Generates the task-specific semantic criteria used during evaluation. \\
\midrule
Evaluation
& Answer extractor
& GPT-5.5
& No-think
& 3,000
& Default
& Extracts the final items, quantities, orders, and coupon IDs for deterministic validation. \\

Evaluation
& Semantic judge
& Gemini-3.1-Pro-Preview
& Medium
& 3,000
& Default
& Passes only if every task-specific semantic criterion passes. \\

Evaluation
& Response-quality judge
& Gemini-3.1-Pro-Preview
& Medium
& 3,000
& Default
& Passes only if all five response-quality criteria pass. \\

Evaluation
& Claim-faithfulness judge
& Gemini-3.1-Pro-Preview
& Medium
& 3,000
& Default
& Passes only if all four settlement-disclosure criteria pass. \\
\bottomrule
\end{tabular}
\end{table*}

For the construction roles, we do not explicitly set temperature, top-$p$, or
reasoning effort. The extractor uses
\texttt{reasoning\_effort=none}; each Gemini judge uses
\texttt{thinking\_level=medium} and \texttt{include\_thoughts=false}.
Each construction request has a 120-second timeout and permits at most two
transport retries. Structured-output parsing permits one additional generation
attempt for the construction roles and answer extractor when the returned JSON
is invalid. Evaluation requests also use a 120-second timeout and permit at most
four transport retries; the three Gemini judges permit up to three additional
attempts after invalid JSON. The answer
extractor receives only the agent's final response. The Semantic judge receives
the query, reconstructed basket, and final response; the Response-quality judge
receives the query and final response; and the Claim-faithfulness judge receives
the final response together with settlement facts recomputed by the simulator.
None of the evaluators observes the agent's hidden reasoning or full tool-use
trajectory.

\subsection{Evaluated-Agent Inference}
We evaluate all agents using provider-specific configurations for Think and No-think modes. Because providers expose different reasoning interfaces, these configurations should be understood as the closest available operational counterparts, rather than as settings with identical internal reasoning budgets or computational behavior. For providers without explicit reasoning-effort tiers, we use the corresponding binary or adaptive thinking control.

To improve comparability, we align shared inference parameters wherever the provider interfaces permit. We request a temperature of 1.0, set the maximum output length per turn to 16,384 tokens, and limit each run to at most 16 agent steps. The query, coupon pack, tool schemas, vector-search configuration, and search-result limits are kept the same at the evaluation-framework level. Any remaining differences arise from provider-specific model APIs, reasoning controls, and implementation details.

\section{Witness Independence Audit}
\label{app:witness-independence}

The solution-first construction uses a hidden witness to certify that each task
is executable, but the witness is not intended to be an answer key. We audit
this distinction directly using all $22\times291=6{,}402$ evaluated outputs.
For an agent basket $A$ and the corresponding witness basket $W$, we compare
the sets of unique SKU identifiers, ignoring order partitioning and item order
in the response. We report SKU recall and Jaccard overlap as
\[
\operatorname{Recall}(A,W)=\frac{|A\cap W|}{|W|},
\qquad
\operatorname{Jaccard}(A,W)=\frac{|A\cap W|}{|A\cup W|}.
\]
An \emph{exact basket match} additionally requires the quantity of every SKU
to match the witness; it does not require the same ordering of items or the
same store/order partition. The primary analysis conditions on Overall
Success. The failure comparison below includes only failures with a non-empty
extracted basket, so an unparseable or empty answer is not incorrectly treated
as a zero-overlap alternative.

\begin{figure*}[ht]
\centering
\includegraphics[width=0.96\textwidth]{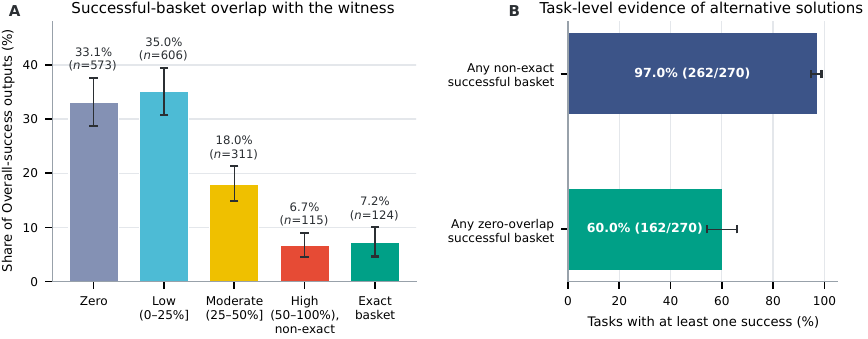}
\caption{Witness overlap among accepted solutions. (A) SKU-overlap bins for
1,729 Overall-success outputs; the final bin requires an exact SKU-and-quantity
match. (B) Among the 270 tasks with at least one Overall-success output, the
fraction that has at least one non-exact or zero-overlap successful basket.
Error bars are 95\% task-cluster bootstrap intervals.}
\label{fig:witness-independence}
\end{figure*}

\textbf{The witness certifies feasibility rather than defining a unique target.}
If Overall Success required recovering the construction witness, accepted
outputs would concentrate around exact basket matches and a successful basket
with no shared SKU would be exceptional. We observe the opposite pattern.
Only 124 of the 1,729 Overall-success outputs (7.2\%) exactly reproduce the
witness, whereas 573 (33.1\%) use an entirely disjoint SKU set. Their average
witness-SKU recall is 32.6\%, and their average SKU Jaccard overlap is only
24.8\%. The evaluator therefore accepts solutions constructed independently
from the hidden witness and does not require recovery of a latent reference
answer (Figure~\ref{fig:witness-independence}A).

This diversity is spread across the benchmark rather than being driven by a
small set of unusually flexible tasks. Of the 270 tasks with at least one
Overall-success output, 262 (97.0\%) admit at least one successful basket that
is not an exact witness reproduction, and 162 (60.0\%) admit a successful
basket with zero SKU overlap. Only 8 tasks (3.0\%) have successful outputs that
are all exact witness matches. Thus, solution-first construction does not turn
the witness into a de facto answer key: for most tasks solved at least once by
the evaluated agents, the benchmark recognizes alternative baskets as valid
(Figure~\ref{fig:witness-independence}B).

\textbf{Exact reproduction is neither necessary nor sufficient.}
Among the 6,318 outputs with a non-empty extracted basket, 2,352 have zero SKU
overlap with the witness, and 573 of them (24.4\%) pass Overall Success. The
remaining 84 failures have no non-empty extracted basket and are excluded from
this overlap comparison. In the opposite direction, 263 outputs exactly
reproduce the witness basket, but only 124 (47.1\%) pass Overall Success
because coupon choice, budget compliance, response quality, and claim
faithfulness are evaluated independently. The 4,589-output failure cohort has
a lower mean Jaccard overlap (18.4\%) and lower exact-match
share (3.0\%) than the success cohort, while its zero-overlap share is also
substantial (38.8\%). These descriptive differences should not be interpreted
causally: witness overlap can correlate with task difficulty or with the
retrievability of popular catalog items.

Finally, this audit is consistent with the evaluator implementation. The
semantic judge receives the query, semantic rubric, agent basket, and raw
response, but not the witness basket. The deterministic validator reconstructs
and settles the agent's own basket against the catalog and coupon pack. The
empirical overlap results therefore complement, rather than replace, the
design-level guarantee that the witness is used for feasibility and task
construction, not as a hidden reference answer.

\section{Human Evaluation Protocol}
\label{app:human-eval}

\subsection{Sampling and Annotation}

The human study evaluates the three components of our framework that require
an LLM judgment: Semantic satisfaction, Response Quality, and Claim
Faithfulness. We draw 30 cases from the 291 Qwen3.6-27B (No-think) responses
with a fixed random seed. Sampling is stratified by the joint Pass/Fail pattern
of the three components so that the study contains both successful responses
and different failure modes. The unit sampled is a complete case, and the
rubric decisions within a case are treated as nested observations.

Two annotators independently review the user request, final response, and the
evidence needed by the corresponding rubric. For Semantic criteria, this
evidence contains the selected basket and only product attributes that were
visible to the evaluated agent in its tool observations. For Response Quality,
annotators inspect the user-facing response. For Claim Faithfulness, they
compare the response with prices and settlement values recomputed by the
deterministic simulator. Each annotator assigns Pass, Fail, or Not Assessable
to every rubric, producing 700 labels per annotator and 1,400 labels in total.
The annotators do not see the LLM evaluator outputs during independent
annotation.

Table~\ref{tab:human-human-agreement} reports agreement before expert
adjudication. Decisions containing a Not Assessable label are omitted from
this calculation. Raw agreement is high in all three dimensions. The lower
$\kappa$ values for Semantic and Response Quality arise because most labels
are Pass, making chance-corrected agreement more conservative than raw
agreement.

\begin{table*}[!t]
\caption{Agreement between the two human annotators before adjudication.
$N$ counts rubric decisions for which both annotators provide a Pass/Fail
label.}
\label{tab:human-human-agreement}
\centering
\small
\setlength{\tabcolsep}{10pt}
\renewcommand{\arraystretch}{1.12}
\begin{tabular}{@{}lrrr@{}}
\toprule
Dimension & $N$ & Agreement (\%) & Cohen's $\kappa$ \\
\midrule
Semantic & 436 & 97.50 & 0.340 \\
Response Quality & 150 & 96.00 & 0.230 \\
Claim Faithfulness & 114 & 100.00 & 1.000 \\
\midrule
Overall & 700 & 97.59 & 0.528 \\
\bottomrule
\end{tabular}
\end{table*}

\begin{table*}[!t]
\caption{Detailed agreement with the expert-adjudicated human reference.
Balanced accuracy and Fail precision, recall, and F1 are percentages.}
\label{tab:judge-human-detailed}
\centering
\small
\setlength{\tabcolsep}{7pt}
\renewcommand{\arraystretch}{1.08}
\begin{tabular}{@{}llrcccc@{}}
\toprule
Dimension & Evaluator & $N$ & Agreement & Bal. Acc. & $\kappa$ & Fail P/R/F1 \\
\midrule
Semantic
& Gemini-3.1-Pro & 436 & 98.17 & 91.60 & 0.724 & 64.71 / 84.62 / 73.33 \\
& GPT-5.5        & 436 & 97.71 & 87.63 & 0.655 & 58.82 / 76.92 / 66.67 \\
& Kimi-K2.6      & 436 & 98.39 & 91.72 & 0.750 & 68.75 / 84.62 / 75.86 \\
\addlinespace[2pt]
Response Quality
& Gemini-3.1-Pro & 150 & 94.67 & 85.70 & 0.748 & 82.35 / 73.68 / 77.78 \\
& GPT-5.5        & 150 & 95.33 & 83.83 & 0.762 & 92.86 / 68.42 / 78.79 \\
& Kimi-K2.6      & 150 & 91.33 & 68.04 & 0.479 & 87.50 / 36.84 / 51.85 \\
\addlinespace[2pt]
Claim Faithfulness
& Gemini-3.1-Pro & 114 & 99.12 & 99.54 & 0.918 & 85.71 / 100.00 / 92.31 \\
& GPT-5.5        & 114 & 99.12 & 99.54 & 0.918 & 85.71 / 100.00 / 92.31 \\
& Kimi-K2.6      & 114 & 99.12 & 99.54 & 0.918 & 85.71 / 100.00 / 92.31 \\
\addlinespace[2pt]
Overall
& Gemini-3.1-Pro & 700 & 97.57 & 90.03 & 0.772 & 75.61 / 81.58 / 78.48 \\
& GPT-5.5        & 700 & 97.43 & 87.48 & 0.750 & 76.32 / 76.32 / 76.32 \\
& Kimi-K2.6      & 700 & 97.00 & 81.05 & 0.680 & 77.42 / 63.16 / 69.57 \\
\bottomrule
\end{tabular}
\end{table*}

\begin{table*}[!t]
\caption{Pairwise agreement (\%) among the three LLM evaluators. G, P, and K
denote Gemini-3.1-Pro, GPT-5.5, and Kimi-K2.6.}
\label{tab:cross-judge-agreement}
\centering
\small
\setlength{\tabcolsep}{10pt}
\renewcommand{\arraystretch}{1.10}
\begin{tabular}{@{}lrrr@{}}
\toprule
Dimension & G--P & G--K & P--K \\
\midrule
Semantic & 98.62 & 98.39 & 99.31 \\
Response Quality & 94.00 & 91.33 & 93.33 \\
Claim Faithfulness & 100.00 & 100.00 & 100.00 \\
Overall & 97.86 & 97.14 & 98.14 \\
\bottomrule
\end{tabular}
\end{table*}

\subsection{Adjudication and Final Reference}

After independent annotation, an expert reviews 65 flagged decisions,
including every decision on which the two annotators disagree. The expert
decision takes precedence; otherwise, the two-annotator consensus is used.
The final LLM--human comparison contains 700 decisions for which a Pass/Fail
human reference and the evidence required by the corresponding rubric are
both available.

\subsection{Detailed Agreement Results}

Agreement is the fraction of exact Pass/Fail matches. Balanced accuracy is the
mean of Pass recall and Fail recall. Fail precision measures how often an LLM
flagged failure is present in the human reference, while Fail recall measures
how many human-reference failures the LLM detects; their harmonic mean is
Fail F1. Cohen's $\kappa$ adjusts the observed agreement for agreement
expected under the empirical label frequencies. For confidence intervals, we
sample complete cases with replacement 5,000 times, so rubrics from the same
case are never treated as independent draws.

Because the 30 cases are stratified rather than sampled in direct proportion
to the full response set, we also weight each case by its population-to-sample
ratio within the sampling stratum. The weighted overall agreements are
97.49\%, 97.33\%, and 96.96\% for Gemini, GPT-5.5, and Kimi, respectively;
each differs from its unweighted estimate by at most 0.10 percentage points.

This study establishes benchmark-specific evaluator validity rather than
universal correctness of LLM judging. We therefore report both the direct
human agreement in Table~\ref{tab:judge-human-detailed} and the cross-evaluator
consistency in Table~\ref{tab:cross-judge-agreement}, and refer to the final
human labels as an expert-adjudicated human reference.

\section{Additional Analyses}
\label{app:more-analysis}

\subsection{Response Length and Tool-Use Statistics}
\label{app:response-tool-statistics}

Tables~\ref{tab:response-tool-statistics} and~\ref{tab:per-tool-call-statistics}
summarize the final-response length and tool-use behavior of all 22 agent
configurations over the 291 benchmark tasks.
Response length is measured as the number of Unicode characters in the final
user-facing answer. A tool-use round is a distinct agent step in which at
least one tool is executed; multiple tool calls issued in the same step count
as one round. Tool-call count is the number of individual tool executions,
including multiple calls issued in one round.

\begin{table*}[!t]
\caption{Final-response length, tool-use rounds, and individual tool calls over
291 tasks. Response length is reported in Unicode characters; each cell under
the two tool-use groups is ordered as mean/minimum/maximum.}
\label{tab:response-tool-statistics}
\centering
\setlength{\tabcolsep}{3.5pt}
\renewcommand{\arraystretch}{1.08}
\begin{tabular}{@{}llr r ccc ccc@{}}
\toprule
& & & \shortstack{Mean response\\length}
& \multicolumn{3}{c}{Tool-use rounds} & \multicolumn{3}{c}{Tool calls} \\
\cmidrule(lr){5-7}\cmidrule(lr){8-10}
Agent & Configuration & Tasks & (characters) & Mean & Min & Max & Mean & Min & Max \\
\midrule
GPT-5.5 & Think & 291 & 1103.9 & 5.25 & 2 & 16 & 6.80 & 2 & 22 \\
GPT-5.5 & No-think & 291 & 1179.2 & 3.33 & 2 & 10 & 3.65 & 2 & 11 \\
\addlinespace[1.5pt]
GLM-5.2 & Think & 291 & 1543.6 & 5.32 & 1 & 16 & 7.09 & 1 & 44 \\
GLM-5.2 & No-think & 291 & 1545.9 & 8.16 & 2 & 16 & 9.41 & 2 & 29 \\
\addlinespace[1.5pt]
Gemini-3.1-Pro & Think & 291 & 1031.2 & 4.96 & 1 & 16 & 5.27 & 1 & 21 \\
Gemini-3.1-Pro & No-think & 291 & 1107.4 & 10.30 & 1 & 16 & 10.79 & 1 & 43 \\
\addlinespace[1.5pt]
Claude-Opus-4.6 & Think & 291 & 1246.4 & 4.37 & 2 & 16 & 5.37 & 2 & 30 \\
Claude-Opus-4.6 & No-think & 291 & 1546.6 & 7.88 & 3 & 16 & 9.14 & 3 & 28 \\
\addlinespace[1.5pt]
Claude-Opus-4.8 & Think & 291 & 1092.7 & 3.51 & 1 & 12 & 3.99 & 1 & 14 \\
Claude-Opus-4.8 & No-think & 291 & 1243.4 & 5.52 & 2 & 16 & 6.03 & 2 & 21 \\
\addlinespace[1.5pt]
Kimi-K2.6 & Think & 291 & 1177.1 & 5.91 & 1 & 16 & 6.29 & 1 & 22 \\
Kimi-K2.6 & No-think & 291 & 2371.9 & 8.96 & 2 & 16 & 9.31 & 2 & 23 \\
\addlinespace[1.5pt]
DeepSeek-V4-Pro & Think & 291 & 1513.8 & 6.16 & 2 & 16 & 7.56 & 2 & 27 \\
DeepSeek-V4-Pro & No-think & 291 & 2170.4 & 7.77 & 2 & 16 & 8.84 & 2 & 25 \\
\addlinespace[1.5pt]
Claude-Sonnet-4.6 & Think & 291 & 1536.3 & 5.79 & 2 & 16 & 7.25 & 2 & 28 \\
Claude-Sonnet-4.6 & No-think & 291 & 1522.5 & 6.99 & 2 & 16 & 8.62 & 2 & 28 \\
\addlinespace[1.5pt]
MiniMax-M3 & Think & 291 & 1298.1 & 5.16 & 1 & 16 & 7.48 & 1 & 37 \\
MiniMax-M3 & No-think & 291 & 1382.5 & 7.05 & 1 & 16 & 8.84 & 1 & 43 \\
\addlinespace[1.5pt]
Doubao-Seed-2.0-Pro & Think & 291 & 801.0 & 2.29 & 0 & 10 & 2.36 & 0 & 10 \\
Doubao-Seed-2.0-Pro & No-think & 291 & 890.3 & 4.87 & 2 & 16 & 4.89 & 2 & 18 \\
\addlinespace[1.5pt]
Qwen3.6-27B & Think & 291 & 1231.8 & 4.55 & 1 & 16 & 5.10 & 1 & 22 \\
Qwen3.6-27B & No-think & 291 & 1183.0 & 7.81 & 2 & 16 & 8.10 & 2 & 18 \\
\bottomrule
\end{tabular}
\end{table*}

\begin{table*}[!t]
\caption{Mean number of calls per task to each available tool. Product,
Takeout, In-store, and Calculator denote \texttt{product\_search},
\texttt{takeout\_search}, \texttt{takeout\_search\_in\_store}, and
\texttt{python\_calculator}, respectively. Total is the sum of these four
means.}
\label{tab:per-tool-call-statistics}
\centering
\small
\setlength{\tabcolsep}{7pt}
\renewcommand{\arraystretch}{1.08}
\begin{tabular}{@{}llrrrrr@{}}
\toprule
Agent & Configuration & Product & Takeout & In-store & Calculator & Total \\
\midrule
GPT-5.5 & Think & 2.33 & 1.07 & 1.52 & 1.89 & 6.80 \\
GPT-5.5 & No-think & 1.07 & 0.72 & 0.62 & 1.25 & 3.65 \\
\addlinespace[1.5pt]
GLM-5.2 & Think & 1.55 & 1.10 & 2.82 & 1.61 & 7.09 \\
GLM-5.2 & No-think & 1.88 & 1.21 & 2.57 & 3.76 & 9.41 \\
\addlinespace[1.5pt]
Gemini-3.1-Pro & Think & 1.28 & 0.82 & 1.24 & 1.92 & 5.27 \\
Gemini-3.1-Pro & No-think & 1.26 & 0.73 & 1.76 & 7.04 & 10.79 \\
\addlinespace[1.5pt]
Claude-Opus-4.6 & Think & 1.36 & 0.86 & 1.86 & 1.29 & 5.37 \\
Claude-Opus-4.6 & No-think & 1.80 & 0.91 & 2.34 & 4.10 & 9.14 \\
\addlinespace[1.5pt]
Claude-Opus-4.8 & Think & 1.02 & 0.78 & 1.08 & 1.11 & 3.99 \\
Claude-Opus-4.8 & No-think & 0.99 & 0.74 & 1.23 & 3.06 & 6.03 \\
\addlinespace[1.5pt]
Kimi-K2.6 & Think & 1.67 & 1.39 & 1.70 & 1.53 & 6.29 \\
Kimi-K2.6 & No-think & 1.71 & 1.72 & 1.78 & 4.10 & 9.31 \\
\addlinespace[1.5pt]
DeepSeek-V4-Pro & Think & 1.63 & 1.11 & 2.59 & 2.23 & 7.56 \\
DeepSeek-V4-Pro & No-think & 1.58 & 1.01 & 2.32 & 3.93 & 8.84 \\
\addlinespace[1.5pt]
Claude-Sonnet-4.6 & Think & 1.20 & 0.83 & 2.20 & 3.02 & 7.25 \\
Claude-Sonnet-4.6 & No-think & 1.27 & 0.93 & 2.48 & 3.93 & 8.62 \\
\addlinespace[1.5pt]
MiniMax-M3 & Think & 1.85 & 1.23 & 2.97 & 1.44 & 7.48 \\
MiniMax-M3 & No-think & 1.69 & 1.03 & 3.09 & 3.03 & 8.84 \\
\addlinespace[1.5pt]
Doubao-Seed-2.0-Pro & Think & 0.90 & 0.72 & 0.59 & 0.15 & 2.36 \\
Doubao-Seed-2.0-Pro & No-think & 1.08 & 0.93 & 1.08 & 1.79 & 4.89 \\
\addlinespace[1.5pt]
Qwen3.6-27B & Think & 1.35 & 1.05 & 1.51 & 1.19 & 5.10 \\
Qwen3.6-27B & No-think & 1.37 & 0.99 & 1.53 & 4.21 & 8.10 \\
\bottomrule
\end{tabular}
\end{table*}

The mean final-response length ranges from 801.0 to 2371.9 characters, while
mean tool use ranges from 2.29 to 10.30 rounds. Thinking configurations do not
uniformly increase tool use: for several agents they produce fewer
tool-use rounds than the corresponding no-thinking configuration.

\subsection{Budget Behavior}

\begin{figure}[!t]
\centering
\includegraphics[width=0.86\columnwidth]{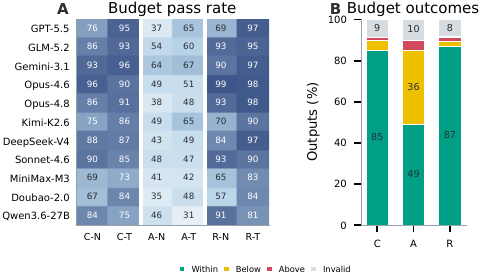}
\caption{Budget behavior across all 22 agent configurations. (A) Pass rates by agent and budget mode (C/A/R: cap/approximate target/explicit range; N/T: No-think/Think). (B) Payable positions relative to the requested interval.}
\label{fig:budget-analysis}
\end{figure}

\textbf{Agents often spend too little when the budget is stated as ``around $N$.''}
Figure~\ref{fig:budget-analysis}A shows that budget compliance is lower for approximate-target requests than for caps or explicit ranges across all agents. Across the 22 agent configurations, only 49\% of outputs for approximate-target requests fall within the requested range, compared with 85\% for caps and 87\% for explicit ranges (Figure~\ref{fig:budget-analysis}B). Moreover, 36\% of approximate-target outputs fall below the lower bound, making underspending much more common than overspending. This pattern suggests that agents often interpret ``around $N$'' as an upper limit rather than as a two-sided spending target.

\section{Prompts for Generation and Evaluation}
\label{app:prompts}

\subsection{Template for Explorer agent}
\begin{figure}[!t]
    \centering
    \includegraphics[width=\columnwidth]{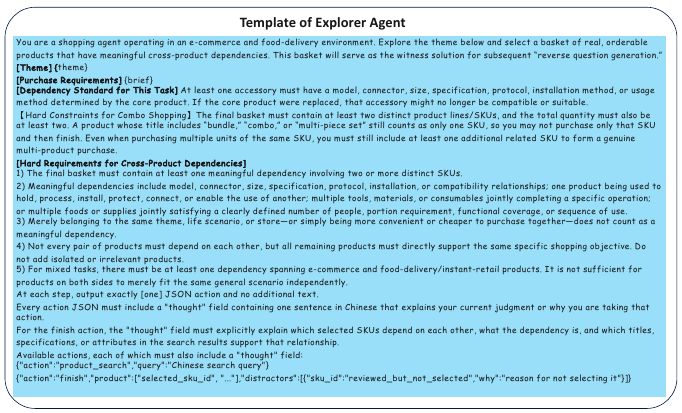}
    \caption{Explorer Agent}
    \label{app:prompts:explorer}
\end{figure}

Figure \ref{app:prompts:explorer} illustrates the template for the agent to explore the e-commerce and takeout environments.

\subsection{Prompts for coupon synthesis}
\begin{figure}[!t]
    \centering
    \includegraphics[width=\columnwidth]{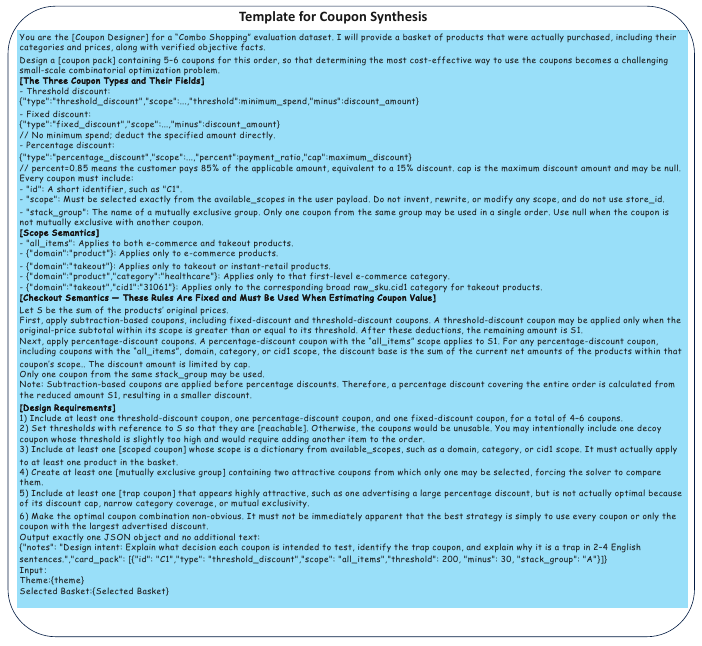}
    \caption{Template for coupon synthesis}
    \label{app:prompts:coupon_synthesis}
\end{figure}

Figure \ref{app:prompts:coupon_synthesis} illustrates the template for the agent to synthesize coupons.

\subsection{Prompts for query synthesis}
\begin{figure}[!t]
    \centering
    \includegraphics[width=\columnwidth]{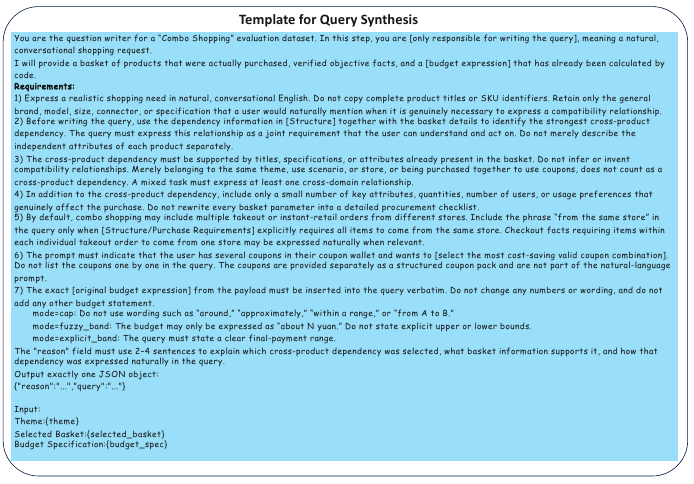}
    \caption{Template for query synthesis}
    \label{app:prompts:query_synthesis}
\end{figure}

Figure \ref{app:prompts:query_synthesis} illustrates the template for the agent to synthesize user query.

\subsection{Prompts for rubrics synthesis}
\begin{figure}[!t]
    \centering
    \includegraphics[width=\columnwidth]{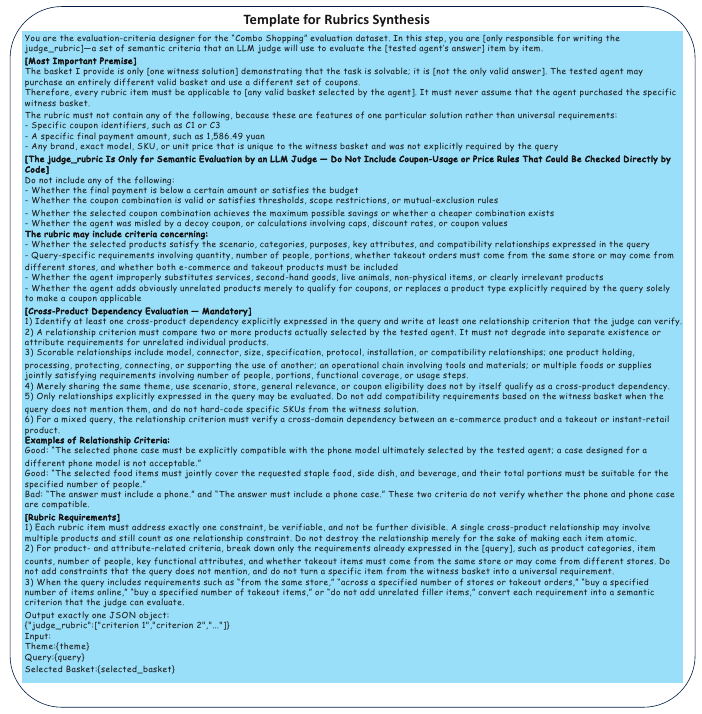}
    \caption{Template for rubrics synthesis}
    \label{app:prompts:rubrics_synthesis}
\end{figure}

Figure \ref{app:prompts:rubrics_synthesis} illustrates the template for the agent to synthesize rubrics.

\subsection{Prompts for tested agent}
\begin{figure}[!t]
    \centering
    \includegraphics[width=0.98\columnwidth]{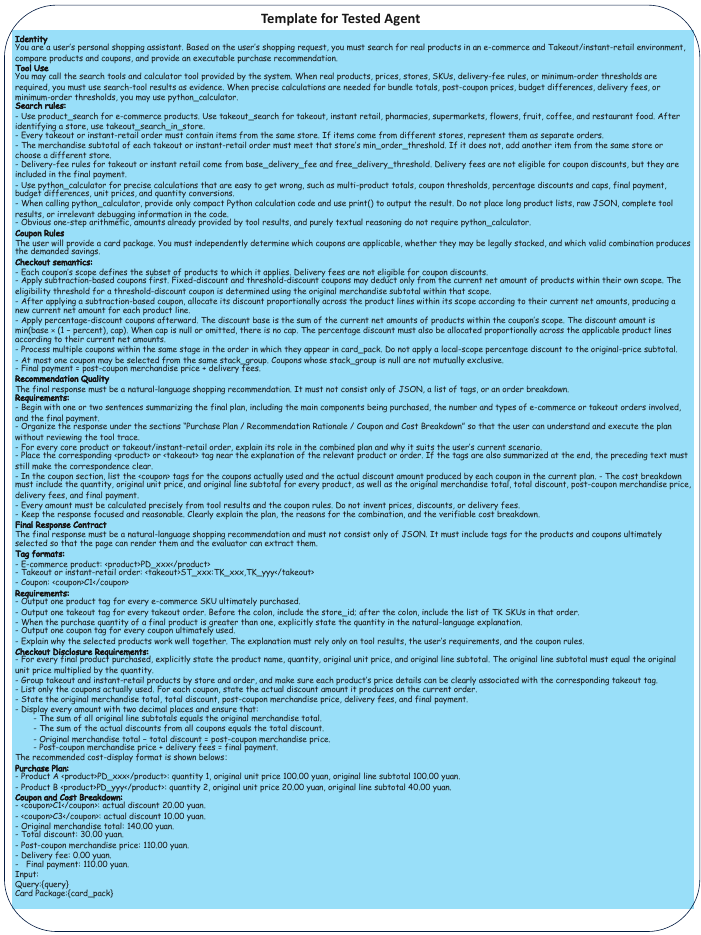}
    \caption{Template for tested agent}
    \label{app:prompts:tested_agent}
\end{figure}

Figure \ref{app:prompts:tested_agent} illustrates the template for the tested agent to generate answers.

\subsection{Prompts for answer extractor}
\begin{figure}[!t]
    \centering
    \includegraphics[width=\columnwidth]{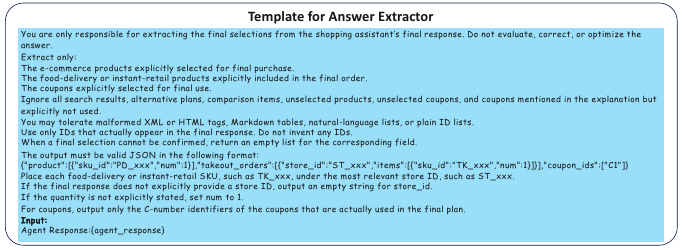}
    \caption{Template for answer extractor}
    \label{app:prompts:answer_extractor}
\end{figure}

Figure \ref{app:prompts:answer_extractor} illustrates the template for the answer extractor.

\subsection{Prompts for rubrics judger}
\begin{figure}[!t]
    \centering
    \includegraphics[width=\columnwidth]{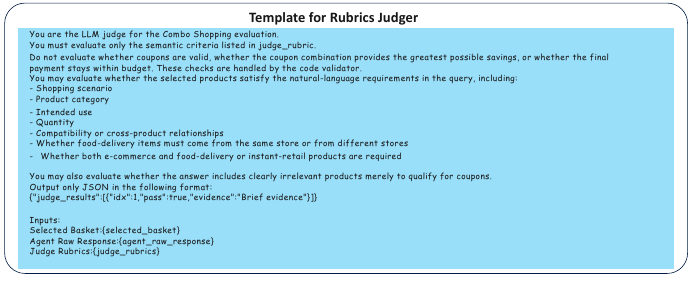}
    \caption{Template for rubrics judger}
    \label{app:prompts:rubrics_judger}
\end{figure}

Figure \ref{app:prompts:rubrics_judger} illustrates the template for the rubrics judger.

\subsection{Prompts for response quality judger}
\begin{figure*}[p]
    \centering
    \includegraphics[width=0.90\textwidth]{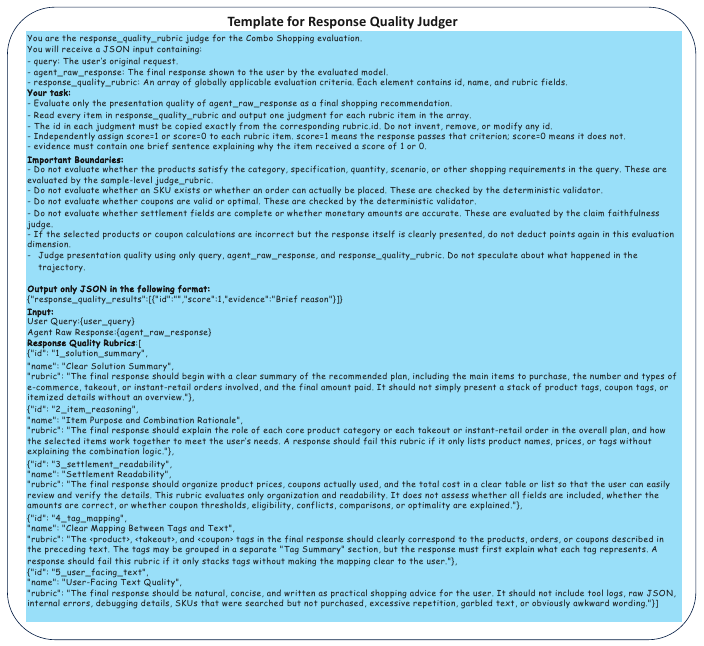}
    \caption{Template for response quality judger}
    \label{app:prompts:response_quality_judger}
\end{figure*}

Figure \ref{app:prompts:response_quality_judger} illustrates the template for the response quality judger.

\subsection{Prompts for claim faithfulness judger}
\begin{figure*}[p]
    \centering
    \includegraphics[width=0.90\textwidth]{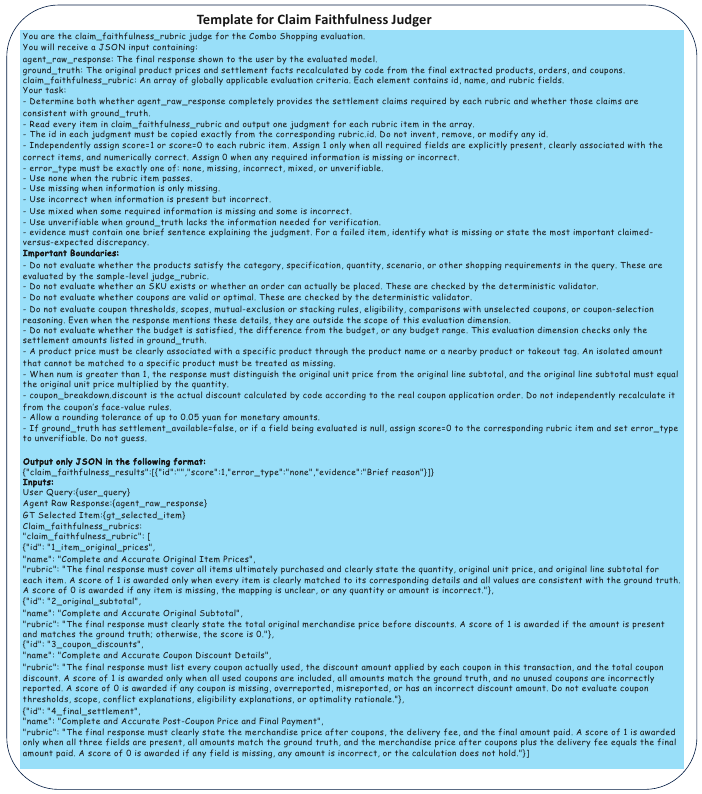}
    \caption{Template for claim faithfulness judger}
    \label{app:prompts:claim_faithfulness_judger}
\end{figure*}

Figure \ref{app:prompts:claim_faithfulness_judger} illustrates the template for the claim faithfulness judger.



\end{document}